\documentclass[runningheads]{llncs}
\usepackage[T1]{fontenc}
\usepackage{graphicx}
\usepackage{enumitem}
\usepackage{multirow}
\usepackage{colortbl}
\usepackage{microtype}
\usepackage{comment}
\usepackage[hang,flushmargin]{footmisc}
\usepackage{booktabs}

\usepackage{amsfonts}
\usepackage{amssymb}
\usepackage{amsmath}
\usepackage[ruled,vlined]{algorithm2e}
\definecolor{commentcolor}{RGB}{110,154,155} % define comment color

\setlist[itemize]{align=parleft,left=0pt,topsep=1mm,itemsep=0mm,parsep=1mm}

\definecolor{azure(colorwheel)}{rgb}{0.0, 0.5, 1.0}
\definecolor{nicegreen}{rgb}{0.0, 0.7, 0.1}
\definecolor{wj}{rgb}{0.01176, 0.5490, 0.5490}
\definecolor{ashblue}{rgb}{0.36, 0.54, 0.66}
\definecolor{ashgrey}{rgb}{0.7, 0.75, 0.71}
\definecolor{applegreen}{rgb}{0.55, 0.71, 0.0}
\definecolor{blue}{rgb}{0.0, 0.0, 1.0}
\definecolor{postechred}{rgb}{0.784, 0.003, 0.313}
\definecolor{gu}{rgb}{0.5460, 0.1755, 0.2766}
\definecolor{jy}{rgb}{0.5, 0.1, 0.8}
\definecolor{ballblue}{rgb}{0.13, 0.67, 0.8}
\definecolor{cornellred}{rgb}{0.7, 0.11, 0.11}
\definecolor{darkcyan}{rgb}{0.0, 0.55, 0.55}
\definecolor{CuGray}{gray}{0.9}
\definecolor{airforceblue}{rgb}{0.36, 0.54, 0.66}
\definecolor{rev}{rgb}{0.784, 0.003, 0.313}
\definecolor{
pink}{cmyk}{0, 0.7808, 0.4429, 0.1412}
\definecolor{amethyst}{rgb}{0.6, 0.4, 0.8}
\definecolor{black}{rgb}{0.0, 0.0, 0.0}
\definecolor{tb3_yellow}{rgb}{0.996, 1.0, 0.6}
\definecolor{tb3_orange}{rgb}{0.980, 0.8, 0.604}
\definecolor{tb3_red}{rgb}{0.972, 0.6, 0.6}
\definecolor{dimgray}{rgb}{0.41, 0.41, 0.41}
\definecolor{brickred}{rgb}{0.8, 0.25, 0.33}
\definecolor{bleudefrance}{rgb}{0.19, 0.55, 0.91}
\definecolor{blue(ncs)}{rgb}{0.265, 0.445, 0.765}
\definecolor{blue(ryb)}{rgb}{0.01, 0.28, 1.0}
\definecolor{orange}{rgb}{1.0, 0.49, 0.0}
\definecolor{Gray}{gray}{0.88}
\definecolor{green(ncs)}{rgb}{0.0, 0.62, 0.42}
\definecolor{brightpink}{rgb}{1.0, 0.0, 0.5}
\definecolor{kellygreen}{rgb}{0.3, 0.73, 0.09}

\newcolumntype{g}{>{\columncolor{CuGray}}c}
\newcolumntype{z}{>{\columncolor{CuGray}}l}

\renewcommand{\paragraph}[1]{\vspace{1mm}\noindent\textbf{#1.}\,\,}

\newcommand{\postechred}[1]{\textcolor{postechred}{#1}}

\usepackage{xspace}

\usepackage{diagbox}

\makeatletter
\def\@fnsymbol#1{\ensuremath{\ifcase#1\or *\or \dagger\or \ddagger\or
   \mathsection\or \mathparagraph\or \|\or **\or \dagger\dagger
   \or \ddagger\ddagger \else\@ctrerr\fi}}
\makeatother

\def\onedot{.\@\xspace}
\def\eg{\emph{e.g}\onedot} 
\def\ie{\emph{i.e}\onedot}

\def\etal{\emph{et al}\onedot}

\newcommand{\Fref}[1]{Fig.~\ref{#1}}
\newcommand{\Tref}[1]{Table~\ref{#1}}

\newcommand{\calL}{{\mathcal{L}}}

\newcommand{\be}{\begin{eqnarray}}
\newcommand{\ee}{\end{eqnarray}}
\newcommand{\bee}{\begin{eqnarray*}}
\newcommand{\eee}{\end{eqnarray*}}

\newcommand{\matrixb}{\left[ \begin{array}}
\newcommand{\matrixe}{\end{array} \right]}

\usepackage{amssymb}
\usepackage{pifont}

\usepackage{wrapfig,lipsum,booktabs}

\usepackage{orcidlink}

\usepackage{pifont}
\renewcommand{\thefigure}{S\arabic{figure}}
\renewcommand{\thetable}{S\arabic{table}}

\usepackage{pifont}
\renewcommand{\thefigure}{S\arabic{figure}}
\renewcommand{\thetable}{S\arabic{table}}

\usepackage[ruled,vlined]{algorithm2e}
\definecolor{commentcolor}{RGB}{110,154,155} % define comment color
\newcommand{\PyComment}[1]{\ttfamily\textcolor{commentcolor}{\# #1}}  
\newcommand{\PyCode}[1]{\ttfamily\textcolor{black}{#1}}

\begin{document}
\title{The Devil is in the Details: Simple Remedies for Image-to-LiDAR Representation Learning}
\titlerunning{Simple Remedies for Image-to-LiDAR Representation Learning}
\author{Wonjun Jo\inst{1}\orcidlink{0000-0003-3894-5483} \and
Kwon Byung-Ki\inst{2}\orcidlink{0000-0003-4187-7944} \and
Kim Ji-Yeon\inst{3}\orcidlink{0000-0002-7535-380X} \and
Hawook Jeong\inst{4}\orcidlink{0009-0003-5935-7055} \and \\
Kyungdon Joo\inst{5}\orcidlink{0000-0002-3920-9608} \and
Tae-Hyun Oh\inst{1,2,3,6}\orcidlink{0000-0003-0468-1571}
}
\authorrunning{W. Jo et al.}

\institute{Department of Electrical Engineering, POSTECH, South Korea \and Graduate School of AI, POSTECH, South Korea \and
Department of Convergence IT Engineering, POSTECH, South Korea \and
RideFlux Inc., South Korea \and
Artificial Intelligence Graduate School, UNIST, South Korea \and
Institute for Convergence Research and Education in Advanced Technology, Yonsei University, South Korea \\
\email{\{jo1jun,byungki.kwon,jiyeon.kim,taehyun\}@postech.ac.kr}, \email{hawook@rideflux.com}, \email{kyungdon@unist.ac.kr}
}

% \institute{
% \noindent\textsuperscript{1}Department of Electrical Engineering, POSTECH, South Korea, 
% \textsuperscript{2}Graduate School of AI, POSTECH, South Korea, 
% \textsuperscript{3}Department of Convergence IT Engineering, POSTECH, South Korea, 
% \textsuperscript{4}RideFlux Inc., South Korea, 
% \textsuperscript{5}Artificial Intelligence Graduate School, UNIST, South Korea, 
% \textsuperscript{6}Institute for Convergence Research and Education in Advanced Technology, Yonsei University, South Korea
% }
%
\maketitle              % typeset the header of the contribution

\begin{abstract}
LiDAR is a crucial sensor in autonomous driving, commonly used alongside cameras.
% where systems often equip both camera and LiDAR sensors.
By exploiting this camera-LiDAR setup and recent advances in image representation learning,
% Leveraging such camera-LiDAR sensor configuration and advances in representation learning in the image domain,
prior studies have shown the promising potential of image-to-LiDAR distillation.
% that image-to-LiDAR distillation is a promising approach.
These prior arts focus on the designs of their own losses to effectively distill the pre-trained 2D image representations into a 3D model.
However, the other parts of the designs have been surprisingly unexplored.
% drawn scant attention. 
We find that fundamental design elements, \eg, the LiDAR coordinate system, quantization according to the existing input interface, and data utilization, are more critical than developing loss functions, which have been overlooked in prior works.
In this work, we show that simple fixes to these designs notably outperform existing methods by 16$\%$ in 3D semantic segmentation on the nuScenes dataset and 13$\%$ in 3D object detection on the KITTI dataset in downstream task performance.
We focus on overlooked design choices along the spatial and temporal axes.
Spatially, prior work has used cylindrical coordinate and voxel sizes without considering their side effects yielded with a commonly deployed sparse convolution layer input interface, leading to spatial quantization errors in 3D models.
Temporally, existing work has avoided cumbersome data curation by discarding unsynced data, limiting the use to only the small portion of data that is temporally synced across sensors.
% , which is a small portion of each dataset.
We analyze these effects and propose simple solutions for each overlooked aspect. \postechred{\href{https://sr-i2l.github.io/}{Project page}}
\end{abstract}

% \keywords{First keyword  \and Second keyword \and Another keyword.}
%
%
%

\section{Introduction}
\label{sec:intro}
Understanding 3D scenes with LiDAR is crucial for autonomous driving. 
With the advance of the neural network, the fundamental 3D scene understanding tasks, \eg, 3D semantic segmentation or 3D object detection~\cite{tang2020searching,zhu2021cylindrical,cheng20212}, have shown promising results with the 3D point cloud annotations~\cite{caesar2020nuscenes,sun2020scalability,behley2019semantickitti}.
However, annotating large-scale 3D LiDAR datasets requires time-consuming efforts with intensive labor~\cite{xie2020pointcontrast,zhang2021self}.
As a workaround, by leveraging 
% the success of 
image representation learning, 
\emph{image-to-LiDAR distillation} methods~\cite{sautier2022image,mahmoud2023self,pang2023unsupervised,liu2023segment} have been developed and demonstrated their effectiveness in an annotation-efficient way.
These methods transfer 2D image representation pre-trained on large data to the data-scarce 3D LiDAR domain.
This facilitates learning rich and transferable 3D representations.

\begin{figure*}[t]
    \centering
    \resizebox{1\linewidth}{!}{%
    \begin{tabular}{c}
    \includegraphics[width=1\textwidth]{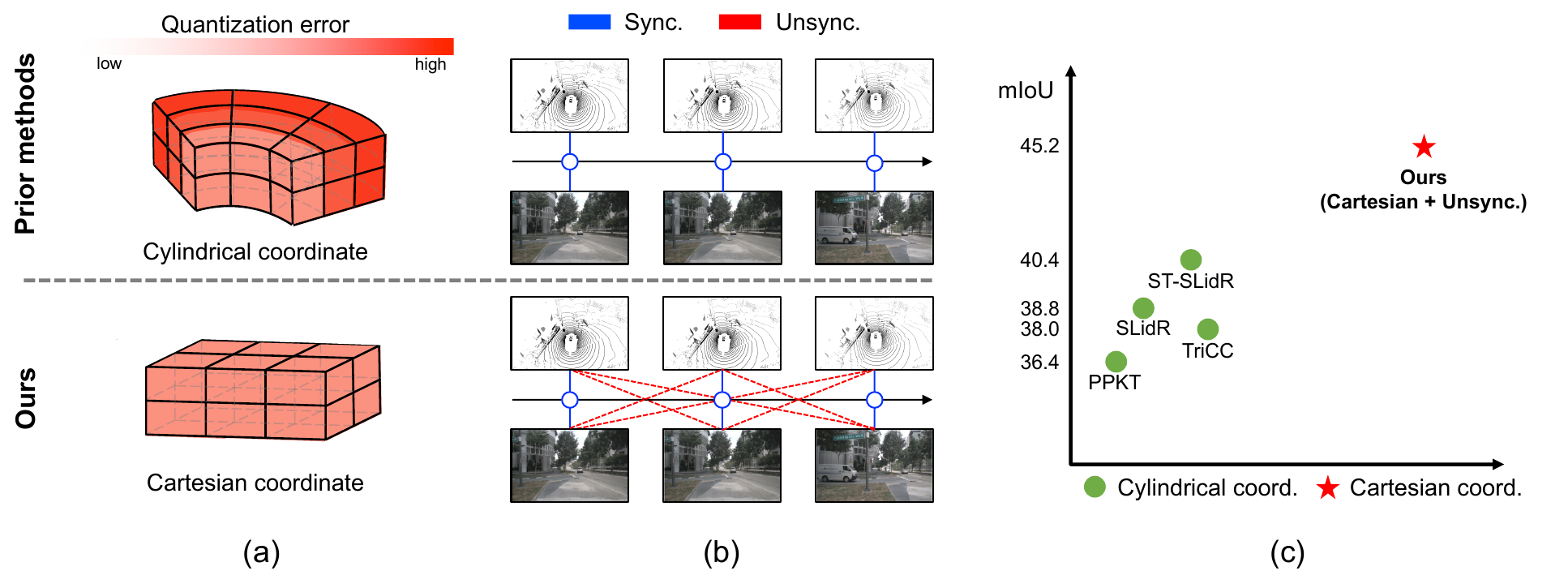}
    \end{tabular}
    }\vspace{-3mm}
    \caption{\textbf{Our simple treatments.}
    (a) The quantization error of cylindrical coordinate increases with distance, while Cartesian coordinate has uniform quantization error regardless of distance. 
    (b) Utilization of unsynced data collected at different times boosts up the combinations of image-LiDAR paired data.
    (c) Compared to the existing methods, 
    we draw attention to underexplored treatments; quantization errors in coordinate system and unsynced data utilization. With these simple treatments, we achieve the state-of-the-art performance in 3D semantic segmentation task. The mIoU is measured by linear probing evaluation protocol in the nuScenes dataset.}
    \label{figure:teaser}
\end{figure*}

A series of recent image-to-LiDAR distillation studies~\cite{mahmoud2023self,pang2023unsupervised,liu2023segment} have focused on improving loss function designs. 
While some have reported good results,
% on challenging benchmarks
they depend very strongly on the standard setting used in SLidR by Sautier~\etal\cite{sautier2022image}, where the SLidR models are used as backbones without modification, and their proposed data processing remains the same.
Following this convention without careful consideration introduces two bottlenecks.

First, previous work has primarily used the cylindrical coordinate as a standard 3D input space for quantizing raw LiDAR point clouds by considering how LiDAR data are acquired.
Zhu~\etal\cite{zhu2021cylindrical} show the effectiveness of the cylindrical coordinate in LiDAR-only tasks, but we found that this is detrimental in our image-to-LiDAR distillation scenario.
The cylindrical coordinate can maintain \emph{uniform density} across varying distances by a trade-off between increasing the voxel size and decreasing the density of LiDAR points according to distance~\cite{zhu2021cylindrical}.
While this appears to be beneficial, when using the standard sparse convolution input layer used in SLidR, the cylindrical coordinate inherently increases quantization errors as the distance increases from the origin~(see Fig.~\ref{figure:teaser}a).
This causes a degradation in the spatial domain, especially at far distances.
Second, recent studies~\cite{sautier2022image,mahmoud2023self,pang2023unsupervised,liu2023segment} 
have only used synced pairs of image-LiDAR data in a dataset to simply guarantee accurate 2D-3D matching without bells and whistles.
However, synced data account for only a small portion of the acquired data and 
discard the majority of the unsynced data.
Thus, a significant portion of temporal information in the existing dataset is under-utilized.

Motivated by these, we propose simple treatments for image-to-LiDAR distillation that effectively handle two unexplored directions: quantization domain and unsynced data.
The first treatment is to change the quantization domain from cylindrical to Cartesian, ensuring \emph{uniform quantization errors} regardless of distances, and to set a much smaller voxel size. 
This simple strategy prevents the degradation of LiDAR points' spatial resolution, ensuring more accurate representation association in image-to-LiDAR distillation.
The second treatment is the use of unsynced data acquired at different times (see Fig.~\ref{figure:teaser}b).  
It should be noted that abundant unsynced data offers additional information for self-supervised training, but accurately estimating point-pixel correspondence may be considered necessary, which is often challenging.
We propose a simple point-pixel matching method called Positive Pair Mining (PPM). 
The PPM module corrects inaccurate point-pixel matching caused by moving objects in a stratified manner, when image and point cloud data are fetched from different times.
This PPM module enables the utilization of numerous unsynched LiDAR-image pair data.

We show that these simple treatments significantly improve the performance in two downstream tasks: 3D semantic segmentation and 3D object detection (see Fig.~\ref{figure:teaser}c).
Our method outperforms all the records the previous state-of-the-art achieved under the benchmark setting.
This hints at several issues with the existing evaluation protocol, \eg, the loose tuning and the dissonance of the design choices of the baselines, which prevents understanding the true performance of the prior work.
This cannot be inferred from each previous work alone. We summarize our main contributions as follows: 
\begin{itemize}
    \item[$\bullet$] 
    We find that prior work underexplored and overlooked fundamental designs, such as the LiDAR coordinate system, quantization, and data utilization.
    Our simple remedies for spatial and temporal designs exhibit significant performance improvements, \ie, state-of-the-art in downstream tasks.
    \item[$\bullet$] 
% We argue that the change of coordinate system from cylinder to Cartesian prevents the degradation of spatial resolution of LiDAR and ensures uniform quantization error. 
    We argue that changing the coordinate from cylinder to Cartesian prevents LiDAR spatial resolution degradation and ensures uniform quantization error.
    \item[$\bullet$] We propose a compact module, \ie, PPM, that corrects incorrect pixel-point matching and enables the model to utilize unsynced image-LiDAR paired data. 
    \item[$\bullet$] We demonstrate consistent improvements, implying that our method not only improves LiDAR representation learning fundamentally but also sets the new baseline and protocol in the field that paves the way for further development. 
\end{itemize}

\section{Related Work}
\label{sec:related}
Our work is related to self-supervised learning that aims to learn useful representations without relying on labeled datasets, which can be categorized based on the data modalities. We briefly review the related lines of work.

\paragraph{2D Image-based Self-Supervised Learning}
Recent approaches have shown remarkable results in learning 2D image representations with 
contrastive learning.
These methods~\cite{chen2020simple,hadsell2006dimensionality,he2020momentum,henaff2020data,misra2020self,oord2018representation,tian2019contrastive,wu2018unsupervised} contrast instance-level representation by augmenting different views of the same instance.
Since contrastive learning is easily affected by the number of negative samples, several methods~\cite{bardes2021vicreg,zbontar2021barlow} propose objective functions that minimize the redundancy of the features and maximize the similarity between the same instance features. 
Such carefully designed contrast-based pre-training regime enables the image representations to be well-transferred to the downstream tasks, \eg, image segmentation or image detection.
As a teacher network, we leverage the powerful pre-trained 2D model, MoCov2~\cite{chen2020improved}, and distill the 2D image knowledge into a 3D student network.

\paragraph{3D Point Cloud-based Self-Supervised Learning}
Analogous to the image domain, 3D self-supervised learning approaches have focused on learning useful representations from the 3D point cloud.
Reconstruction-based methods~\cite{chen2021shape,sauder2019self,wang2021unsupervised} generate object-level paired representations by perturbing point clouds, \eg, distortion, random arrangement, or occlusion in pre-training.
Most recent work~\cite{hou2021exploring,xie2020pointcontrast,nunes2022segcontrast,zhang2021self,yin2022proposalcontrast} allows object-level 3D representations to transfer into 3D object detection or 3D segmentation while utilizing a single modality dataset, a 3D point cloud.
Temporal consistency-based methods~\cite{wu2023spatiotemporal,nunes2023temporal} find the corresponding point cloud across different timestamps by tracking the objects. STRL~\cite{huang2021spatio} exploits spatio-temporal cues from a 3D point cloud and generates temporally-correlated frames to learn invariant representations.
While STRL proposes spatial/temporal data alignments for 3D self-supervised learning, single-modality representations suffer from a lack of appearance information in images. 

\paragraph{2D-to-3D Distillation-based Self-Supervised Learning}
Multi-modal self-supervised learning leverages pre-trained 2D image representations and distills them into 3D representations by applying contrastive loss to maximize the similarity between 2D-3D paired features.
PPKT~\cite{liu2021learning} proposes a pixel-to-point knowledge transfer learning method. They use the back-projection function to align 2D and 3D features and transfer knowledge between heterogeneous networks.
SLidR~\cite{sautier2022image} contrasts 2D image superpixels and corresponding 3D point cloud following SLIC~\cite{achanta2012slic}.
Based on SLidR, ST-SLidR~\cite{mahmoud2023self} proposes the semantically-tolerant and class-agnostic balanced loss.
% \todo{(need to add superpixel reference)}
Due to the heterogeneous data acquisition frequencies of LiDAR and camera sensors, 
the aforementioned methods utilize only \emph{synced}
images-LiDAR 
pairs
for multi-modal contrastive learning.
Recent work, TriCC~\cite{pang2023unsupervised},
propose the triplet loss function enabling the utilization of extended timestamps while they only fetch \emph{synced} frames. 
Moreover, they obtain a pixel-point matching table by computing the similarity between pixel features and point features.
Instead, we directly compute a 3D transformation matrix to obtain accurate correspondence between points and pixels.
Another line of work, Seal~\cite{liu2023segment} replace the superpixel-driven matching~\cite{achanta2012slic} with the foundation models, \eg SAM~\cite{kirillov2023segment}, X-Decoder~\cite{zou2023generalized}, OpenSeeD~\cite{zhang2023simple}, and SEEM~\cite{zou2024segment}.
ScaLR~\cite{puy2024three} improves image-to-LiDAR distillation performance by scaling up the dataset, 2D backbone, and 3D backbone.
Though not 
% 2D-to-3D
distillation, GPC~\cite{pan2023gpc} present 
% is a study on self-supervised learning that 
a point cloud colorization method as a pretext task loss to match synchronized images.
The aforementioned methods focus on the designs of their own losses to effectively distill the pre-trained 2D image representations into the 3D model. Instead, we carefully examine the fundamental designs of the input interface and data utilization, which were overlooked by the prior art.

\section{Simple Treatments}

\begin{figure*}[t]
\centering
\resizebox{1\linewidth}{!}{%
    \begin{tabular}{c}
    \includegraphics[width=1\textwidth]{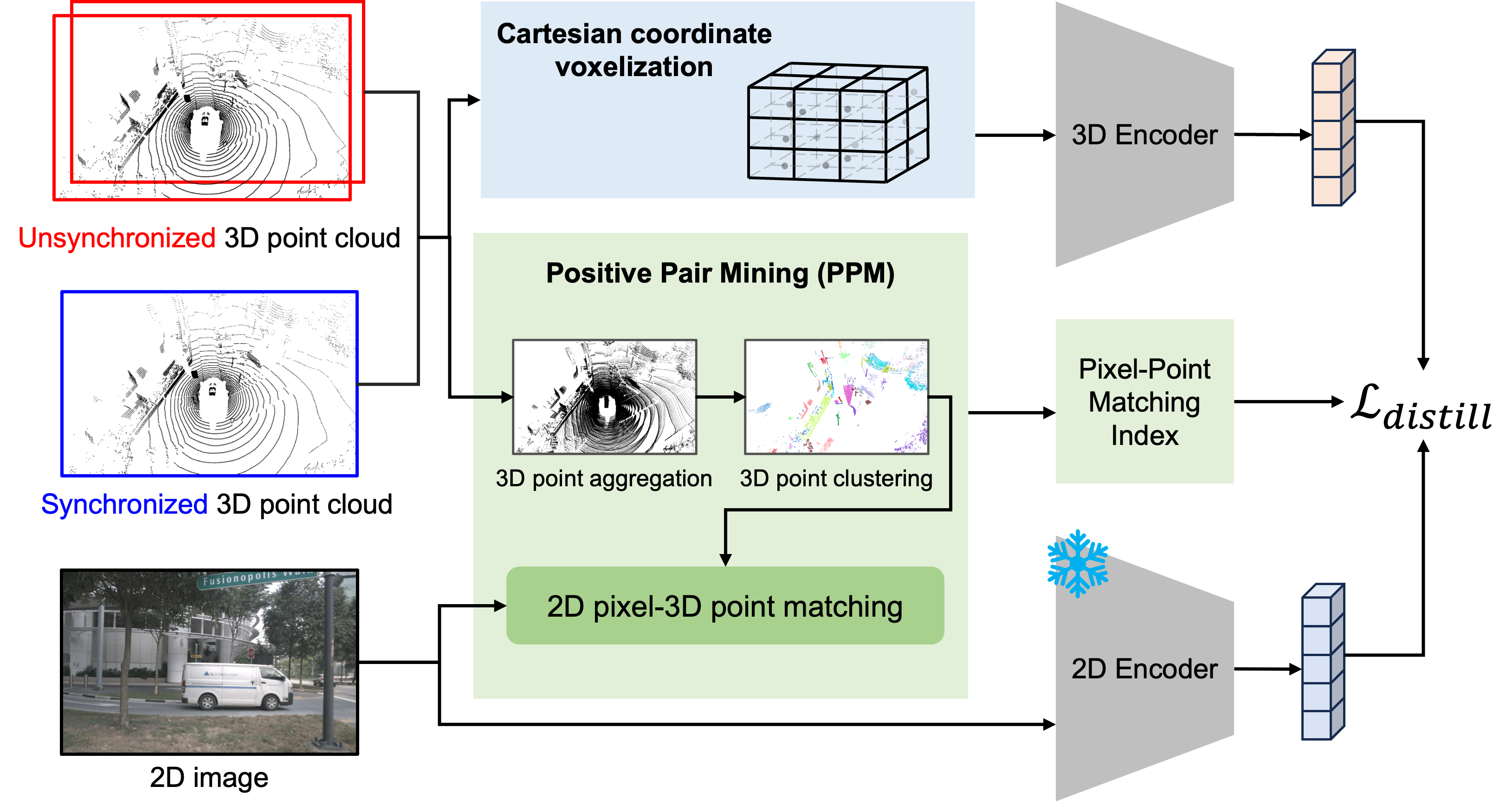}
    \end{tabular}
    }
    \caption{\textbf{Overall pipeline}. Given 3D LiDAR point cloud and 2D image, 
    our goal is to distill pre-trained 2D image representations into a 3D backbone model.
    Based on our observation that Cartesian coordinate prevents degradation in spatial resolution of LiDAR and ensures uniform quantization error, synced/unsynced points are quantized in Cartesian domain and then 3D encoder extracts the point-wise features. Unlike the prior work, we utilize unsynced point cloud for pixel-point matching. Our PPM module corrects inaccurate pixel-point matching for moving objects. Note that our pre-trained 2D encoder is frozen during training. We contrast the matched pixel-point features.}
    \label{figure:pipeline}
\end{figure*}

In this section, we first explore the changing coordinates for proper quantization (Sec. \ref{sec:coordinate}). Next, we provide the details of the PPM module for addressing unsynced data (Sec. \ref{sec:unsynced}). Finally, we outline a pipeline for integrating these treatments into image-to-LiDAR distillation (Sec. \ref{sec:learning}).

\subsection{Treatment 1: Cartesian Coordinate}\label{sec:coordinate}

\begin{figure}[t]
    \centering
    \includegraphics[width=\linewidth]{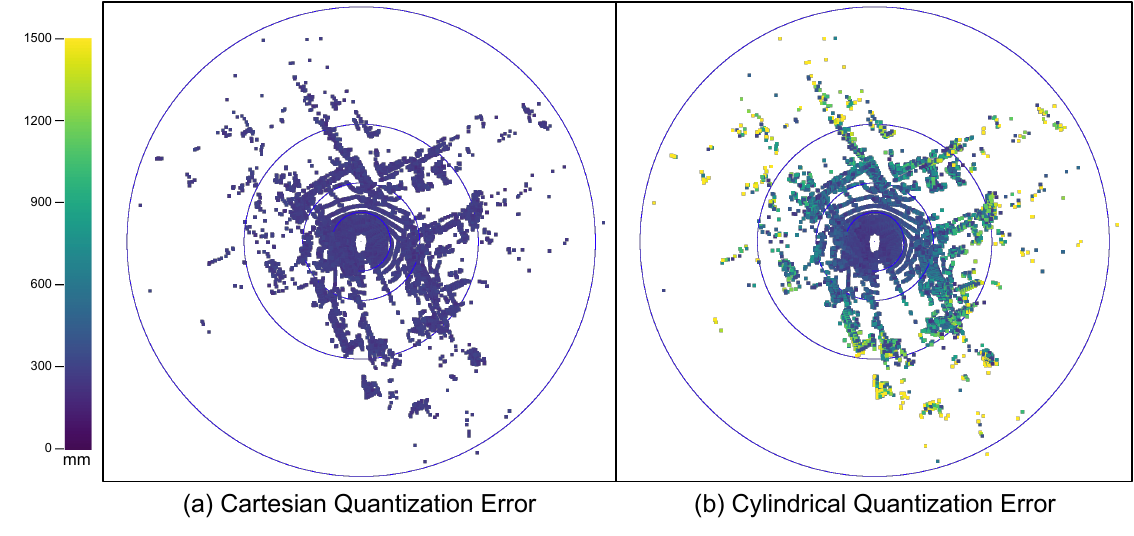}
\caption{
    \textbf{
Quantization error characteristics differ by coordinate system.} 
    (a) In the case of Cartesian quantization, the voxel size remains constant regardless of the distance, resulting in a uniform quantization error that does not vary with distance. On the other hand, (b) with Cylindrical quantization, the voxel size increases as the distance increases, leading to a rise in quantization error due to its characteristic of expanding voxel size with distance.
    }\label{figure:quan_error}
\end{figure}

In the realm of LiDAR sensor data, cylindrical coordinate mitigates the issue of increasing sparsity in LiDAR point clouds with distance~\cite{zhu2021cylindrical}. 
This approach allows for voxel sizes to expand with distance, enhancing the likelihood of points being included within a voxel. 
As a result, the density of voxels remains uniformly distributed, irrespective of the distance from the origin.
However, as the distance increases, so does the voxel size, leading to a larger average distance between the original and quantized positions of points within a voxel. 
This signifies an increase in the quantization error as one moves further from the origin (see Fig. \ref{figure:quan_error}a).
Consequently, with increasing distance from the origin, the enlarged voxel volume results in a lower data resolution, preserving details at close range while significantly losing details at far distances.
This ultimately leads to a degradation in overall data resolution.
Conversely, using Cartesian coordinates maintains the consistent voxel size throughout the space, resulting in the uniform quantization error across all distances. 
The average error determined by voxel size remains constant, regardless of position (see Fig. \ref{figure:quan_error}b). 
Thus, while some information may be lost at close range, the loss of detail at larger distances is considerably lessened.
This ultimately results in the preservation of overall data resolution.

We postulate that small and uniform quantization error is more critical than the emphasis on uniform density~\cite{zhu2021cylindrical} in performance.
The degradation of overall data resolution would confine image-to-LiDAR distillation, causing a decline in performance.
To alleviate this, we propose the adoption of cartesian coordinates in place of cylindrical coordinates within the image-to-LiDAR distillation scheme.
Our first simple treatment involves a modification to the conventional process shared by existing image-to-LiDAR distillation methods before the point cloud feed-forward to the 3D network, which is as follows: 
\begin{enumerate}
    \item Transforming the raw 3D points from cartesian to cylindrical coordinate.
    \item Voxelizing of the input with 3D cylindrical partitioning, with voxel sizes of $\{\delta\rho = 10cm, \delta\phi = 1^{\circ}, \delta z = 10cm\}$ and quantizing 
% for the sparse convolutional layer, \eg, \cite{choy20194d}.
\end{enumerate}
We propose a modification to this process:
\begin{enumerate}
    \item \textbf{Not} transforming the 3D points from cartesian to cylindrical coordinate.
    \item Voxelizing of the input using 3D cartesian partitioning, with voxel sizes of $\{\delta x=10cm, \delta y=10cm, \delta z = 10cm\}$ and quantizing.
\end{enumerate}
By cartesian coordinates, we could alleviate overall resolution degradation, ensuring that image-to-LiDAR distillation is not constrained by degraded resolution. 

\subsection{Treatment 2: Utilization of Unsynced Data}\label{sec:unsynced}

\begin{figure}[t]
    \centering
    \includegraphics[width=\columnwidth]{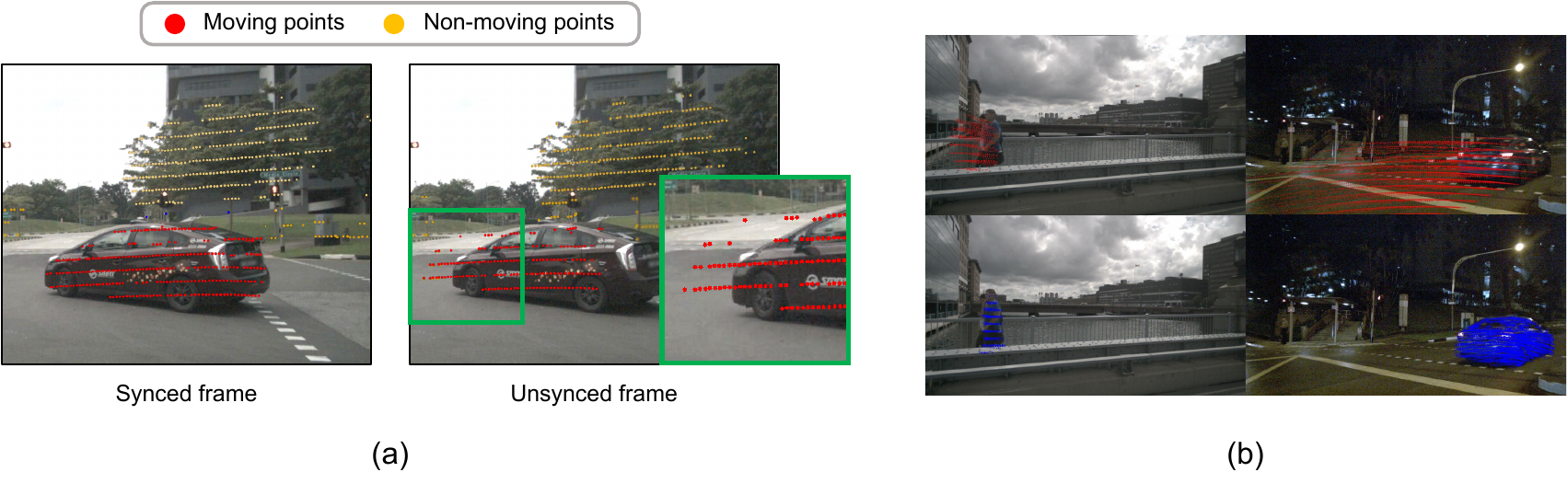}
\caption{
    \textbf{Visualization of the effects of PPM module.}
    (a) Compared to the synced frame, unsynced LiDAR points are misaligned with pixels in moving object regions. 
    (b) With the PPM module, we observe that moving points are well matched with pixels in the corresponding 2D images. 
    }\label{figure:ppm_qual}
\end{figure}

Utilizing only synced data for image-to-LiDAR distillation learning significantly limits data utilization. Intuitively, given $N$ number of subsequent point clouds [$P^{1}$, ..., $P^{t}$, ..., $P^{N}$] and images [$I^{1}$, ..., $I^{t}$, ..., $I^{N}$], where $t$ denotes time, the available data is limited to $N$ instances. 
Thanks to various combinations of LiDAR-image data, we can increase the quantity of data by utilizing unsynced data from different timestamps instead of only using synced data from the same timestamp.
However, additional data processing on moving points is required to leverage unsynced data. Figure~\ref{figure:ppm_qual}a demonstrates the reason. In the unsynced case, the non-moving points are accurately aligned with objects in the 2D images, but the moving points are mismatched because obtaining projection from moving points to the 2D image is intractable. To utilize unsynced data, we propose a simple treatment, \ie, the Positive Pair Mining (PPM) module. 
The PPM module dynamically registers point clouds from the source to the target timestamp, regardless of whether the objects are moving or stationary.
Using the calibration matrix provided with the synced data, the dynamically registered point clouds can be more accurately matched to the images corresponding to the target time.

Specifically, let $P^{t}$
% =[\boldsymbol{p}^{t}_{1}, \cdots, \boldsymbol{p}_{N}]$ 
and $I^{s}$ denotes the point cloud and image, which are acquired at time $t$ and $s$ from different sensors, respectively, which we represent
% The $P^{t}$ and $I^{s}$ correspond to
% \begin{equation}
    $P^{t} = [\boldsymbol{p}^{t}_{1}, ..., \boldsymbol{p}^{t}_{N_{t}}]$ and
    % \quad \textrm{and} \quad 
    $I^{s} \in \mathbb{R}^{H \times W \times 3}$,
% \end{equation}
where $\boldsymbol{p}^{t}_{i} \in \mathbb{R}^{3}$ denotes the $i$-th 3D point at the time $t$ in cartesian coordinate, and $H$, $W$ and $N_{t}$ the height and width of the image, and the number of points at time $t$, respectively.
% $\boldsymbol{p}^{t}_{i} \in \Real^{3}$ and $N$ refer the $i$th 3D point and the number of points. 
Given a known pre-calibrated relative pose between the LiDAR and the camera, the 3D-to-2D projection $T_{t, s}:\mathbb{R}^{3} \rightarrow \mathbb{R}^{2}$ outputs the projected 2D coordinate of $I^{s}$, \ie, 
% that is,
$\mathbf{x}^s_i = T_{t,s}(\boldsymbol{p}^{t}_{i})$. If $t=s$, $T_{t,s}$ ensures the accurate 3D-to-2D correspondence between $\boldsymbol{p}^{t}_{i}$ and  $\mathbf{x}^s_i$. However, for 
% in the case of
$t\neq s$, it does not guarantee the correspondence. To address this potentially unreliable matching case, we design the PPM module. 
Given point clouds at an unsynced frame, 
we first inspect whether they are likely moving because the projection is unreliable with moving 3D points. 
we subsequently find the more accurate 3D-to-2D matching by the mining of a 3D transformation matrix $Z_{t,s,i}{:\,}\mathbb{R}^{3} \rightarrow \mathbb{R}^{3}$ for the moving points. To be concise, we omit the subscript indices. 
$Z$ is combined with $T$ to create the pixel-point matching index.
As shown in Fig.~\ref{figure:ppm_qual}b, moving points are better matched to the image when utilizing the PPM module than when it is not.

Our proposed PPM module consists of unsupervised methods. Notably, it involves aggregating unsynced and synced 3D point clouds from consecutive frames and applying 3D point clustering to detect objects. Additionally, it includes ground removal to separate objects, moving cluster tracking to identify moving objects, and cluster-wise Iterative Closest Point (ICP) to register points for moving objects specifically.
See supplementary material for details.

\subsection{Image-to-LiDAR Distillation with Simple Treatments}\label{sec:learning}

We combined the two simple treatments to construct the overall pipeline. Figure~\ref{figure:pipeline} demonstrates the overall pipeline. Compared to the baseline~\cite{sautier2022image}, we employ cartesian coordinate and enable 3D representation learning using unsynced data by introducing the PPM module.

As shown in~\Fref{figure:pipeline}, we sample the 3D point clouds $P^{t}$ and $P^{s}$, where t and s denote synced and unsynced frames. We also sample the 2D image $I^{t}$ at synced frame. The point clouds quantized in cartesian coordinate and image pass through the 3D and 2D networks, resulting in corresponding 3D point-wise features $\boldsymbol{F}^{t} \in \mathbb{R}^{N'_{t} \times D}$  and $\boldsymbol{F}^{s} \in \mathbb{R}^{N'_{s} \times D}$, and 2D pixel-wise features $\boldsymbol{G}^{t} \in \mathbb{R}^{H \times W \times D}$. 
Because of quantization, $N'$ is lower than $N$ and the number of pixel-point matching index is reduced.
Then, we find the matching between $\boldsymbol{F}$ and $\boldsymbol{G}$ using the pixel-point matching index. At the time $t$, the pair of 3D point and 2D pixel features $\{\mathbf{f}_{i}^{t}, \mathbf{g}_{i}^{t}\}$ are follows: 
\begin{equation}
\{\boldsymbol{f}_{i}^{t}, \boldsymbol{g}(\mathbf{x}^t_i) \mid \mathbf{x}^t_i = T(\boldsymbol{p}_{i}^{t})\},
\end{equation}
where $\boldsymbol{g}$ is function that exploits 2D pixel feature $\mathbf{g}_{i}^{t}$ from $\boldsymbol{G}^{t}$ at 2D coordinate $\mathbf{x}^t_i$. If the 3D point comes from the unsynced frame $s$, we employ $Z$ for the pairing of $\{\mathbf{f}_{j}^{s}, \mathbf{g}_{j}^{t}\}$, which defined as: 
\begin{equation}
\{\boldsymbol{f}_{j}^{s}, \boldsymbol{g}(\mathbf{x}^t_j) \mid \mathbf{x}^t_j = T(Z(\boldsymbol{p}_{j}^{s}))\}.
\end{equation}
Note that $Z$ is the identity matrix when applied to non-moving points, including ground points.
We then gather $M=N'_t+N'_s$ number of paired features $\{\mathbf{f}_{k},\mathbf{g}_{k}\}_{k=1}^{M}$ from the synced and unsynced data. Then, we apply the image-to-LiDAR contrastive distillation loss to enforce the 3D feature to be similar to the paired 2D feature.
We use the same self-supervised loss with SLidR~\cite{sautier2022image} to see the effects of our changes without any fancy loss design, which 
% The self-supervised loss $\calL_{distill}$ for distilling 2D representations 
is defined as
% \begin{equation}
%     \calL = -\frac{1}{M}\sum_{k=1}^M\log[\frac{\exp(\langle\mathbf{f}_k,\mathbf{g}_k\rangle/\tau)}{
%     \sum_{d\neq k }^M{\exp(\langle\mathbf{f}_k,\mathbf{g}_d\rangle/\tau) + \exp(\langle\mathbf{f}_k,\mathbf{g}_k\rangle/\tau)}}],
% \end{equation}
\begin{equation}
    \calL_{distill} = -\frac{1}{M}\sum\nolimits_{k=1}^M\log[\tfrac{\exp(\langle\mathbf{f}_k,\mathbf{g}_k\rangle/\tau)}{\sum_{d=1}^M{\exp(\langle\mathbf{f}_d,\mathbf{g}_k\rangle/\tau)}}],
\end{equation}
where $\langle\cdot, \cdot\rangle$ denotes the normalized inner 
% scalar
product in the $l_2$-normalized $\mathbb{R}^D$, and $\tau > 0$ is the temperature term. 
The positive and negative pairs for the contrastive learning are determined following the implementation of Sautier~\etal~\cite{sautier2022image}, which groups the pixel-wise features by super-pixel segmentation~\cite{achanta2012slic} and their corresponding point-wise features. We simplify the above expressions by using pixel-wise representation instead of superpixel-wise.

\section{Experiments}
\label{sec:experiment}
In this section, we describe the pre-training details of 3D network (Sec.~\ref{sec:pretrain}). 
In Sec.~\ref{sec:ablation}, we present the fine-tuning results with varying amounts of annotation and ablation studies of the proposed PPM module and coordinates.
Then, we provide the experiment results on the downstream tasks (Sec.~\ref{sec:segmentation} and Sec.~\ref{sec:detection}).

% class fine-tune
% object detection
% \input{tables/table11} 

\subsection{Experiment Setup}\label{sec:pretrain}

% The experiment setup largely follows the prior work~\cite{sautier2022image} for fair comparison.

\paragraph{Networks}
The 3D backbone network is composed of sparse residual U-Net architecture~\cite{choy20194d} and a linear layer, which projects 3D features to $D$-dimensional space. The sparse residual U-Net architecture is composed of 
$3 \times 3 \times 3$ with sparse convolutions.
The input to the 3D backbone network is voxelized in either cylindrical or Cartesian coordinates, with voxel sizes altered to 10$cm$ or 5$cm$ for experimentation. This aspect differs from traditional image-to-LiDAR distillation methods.
The 2D network is composed of 1) MoCov2~\cite{chen2020improved} that is a frozen network pre-trained by a self-supervised method, whose architecture is ResNet 50~\cite{he2016deep}, 2) a trainable linear layer that projects 2D features to $D$-dimensional space, and 3) the bilinear upsampling layer to obtain a pixel-wise feature of input resolution from a reduced output resolution.
% The architecture of 2D feature extractor $f$ is ResNet 50~\cite{he2016deep} pre-trained using MoCov2~\cite{chen2020improved}, a linear layer, and a bilinear upsampling layer to obtain a pixel-wise feature of input resolution from a reduced output resolution.

\paragraph{Pre-training Datasets}
We pre-train the 3D network on the nuScenes dataset~\cite{caesar2020nuscenes}, which 
% The dataset
has $700$ training scenes, and split into $600$ remaining training scenes and $100$ mini-validation scenes for choosing the training hyperparameters.
% and . 
% is split into $700$ number of training scenes and $100$ number of mini-validation scenes. 
% All 3D backbones are pre-trained on the nuScenes dataset~\cite{caesar2020nuscenes}.
% This dataset encompasses 700 training scenes among which 100 scenes are set aside to form our mini-validation split. 
% This split is utilized for conducting hyperparameter tuning and evaluation.
% The sampling rates of Lidar and camera are 20Hz and 12Hz, respectively, resulting in synchronized frames, \ie, \textit{key frames}, with a sampling rate of 2Hz. All other frames other than the \textit{key frames} are \textit{unsynced frames}.
The sampling rates of LiDAR and the camera are 20Hz and 12Hz, respectively, and synced data has a sampling rate of 2Hz.
% The frequencies of LiDAR and camera are 20Hz and 12Hz, respectively, of which the frequency of synchronized key frame data is 2Hz, and the rest are unsynchronized sweep frame data. 
We refer to the timestamps of synced data at 2Hz as keyframes and the timestamps of the remaining frame data as inter-frames. The LiDAR and image data belonging to inter-frames are inherently unsynced due to the difference in data collection frequencies.
We train the 3D network with keyframe and inter-frame data using the proposed method.
% The 3D backbones with our approach are pre-trained by utilizing all key frame and sweep frame data derived from the 600 remaining training scenes.

\paragraph{Data augmentation}
We follow the data augmentation strategy used in 
% the SLidR framework
Sautier~\etal~\cite{sautier2022image}. 
For the 3D point clouds, we apply the composition of random transformations, including a z-axis rotation, a 50\% probability random flip against the x and y-axes, and a random cuboid dropping. For the images, we apply the random horizontal flip and random crop-resizing transformations.

\paragraph{Hyperparameters}
% The 3D backbone network is pre-trained 
The 3D backbone network is pre-trained using 4 A100 GPUs, employing a batch size 16 and spanning 50 epochs. We adopt the Stochastic Gradient Descent (SGD) for the optimization algorithm with an initial learning rate set at 0.5, momentum of 0.9, weight decay of 0.0001, and dampening of 0.1. We use the cosine annealing scheduler that progressively reduces the learning rate from its initial value to $0$ by the conclusion of the $50^{th}$ epoch.

\subsection{Ablation Studies} \label{sec:ablation}
\begin{table}[t]
    \caption{\textbf{Impact of coordinate and voxel size.} 
When using cylindrical coordinates, average quantization error is larger, leading to a reduction in data resolution. In contrast, using Cartesian coordinates results in a smaller average quantization error, maintaining data resolution. Reducing voxel size decreases the quantization error and preserves data resolution better. As data resolution is well preserved, performance improves.}
    \centering
    \resizebox{0.8\linewidth}{!}{
    \begin{tabular}{l@{\hskip 0.2in}c@{\hskip 0.2in}c@{\hskip 0.2in}c@{\hskip 0.2in}c}
        \toprule
        \multirow{2}{*}{Method} & \multirow{2}{*}{Coordinate} & \multirow{2}{*}{Voxel size} & \multirow{2}{*}{Quan. error} & nuScenes \\
        \cmidrule(lr){5-5}
        & & ($cm$) & ($mm$) & Lin. Prob. (100$\%$) \\
        
        \midrule
        
        SLidR~\cite{sautier2022image} & Cylindrical & 10 & 229.2 & 38.8 \\
        
        Ours & Cartesian & 10 & 96.5 & 40.8 \\
        
        Ours & Cartesian & 5 & 48.2 &\textbf{41.2} \\
        \bottomrule
    \end{tabular}}
    \label{table:ablation_coord}
\end{table}
\paragraph{Coordinate system and voxel size}
\Tref{table:ablation_coord} ablates the effect of the coordinate system and voxel size.
As shown in~\Fref{figure:quan_error}, cylindrical coordinate increases the spatial quantization errors as the distance increases, which introduces performance degradation to 3D models. We thus convert the input coordinate system from cylindrical to Cartesian, which ensures the uniform quantization error regardless of the distances. This approach reduces the overall quantization error and preserves the data resolution well. Additionally, by decreasing the voxel size, we further reduce the overall quantization error, thereby preserving the data resolution even more. Our method with the Cartesian coordinate interface enjoys a clear benefit of +2.0\% in the 3D semantic segmentation task with nuScenes dataset. 
Moreover, we obtain an additional +0.4\% gain with the smaller voxel sizes. We postulate that preserving data resolution effects improves image-to-LiDAR distillation learning and results in performance improvement.
Therefore, our joint treatment of the spatial aspect, \ie, coordinate change and voxel size adjustment, is more impactful than the prior art that only focuses on loss designs.

\begin{figure*}[t]
\centering
\resizebox{1\linewidth}{!}{%
\begin{tabular}{c}
\includegraphics[width=1\textwidth]{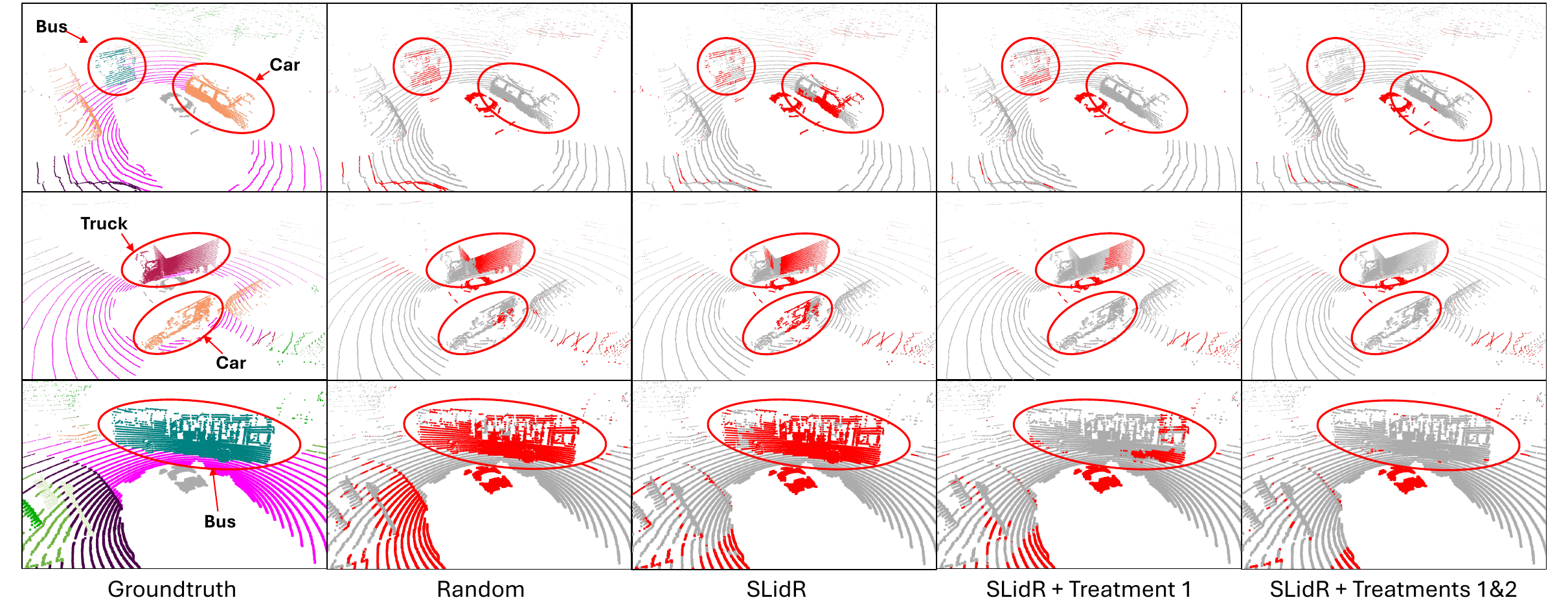}
\end{tabular}
}
\caption{
    \textbf{Qualitative results of error map. 
    } Error map among the Random, SLidR~\cite{sautier2022image}, SLidR w. Cartesian, and Ours when fine-tuned with 1\% labeled data. Gray and red colors denote correct and incorrect points, respectively. We visualize the results of the validation set in nuScenes~\cite{caesar2020nuscenes}. When our treatments were applied sequentially, we observe the qualitative improvement. Treatment1 and treatment2 denote Cartesian coordinate and utilization of unsynced data, respectively. 
    }
    % \vspace{-3mm}
    \label{figure:supp_fig2}
\end{figure*}

\paragraph{Unsynced data utilization}
In this ablation study, the cartesian coordinate system and a voxel size of 5$cm$ were applied as the default settings. In~\Tref{table:ablation_ppm}, we evaluate the impact of our utilization of unsynced frames and our proposed Positive Pair Matching(PPM) module.
Since unsynced data is collected at different times,
the misalignments between 2D pixels and 3D points occur in moving objects. 
Without dedicated adjustment, the sole use of unsynced frames hinders the distillation and causes performance degradation.
Our proposed PPM module resolves the issue by correcting inaccurate pixel-point matching and enables the utilization of numerous unsynched data pairs.
Our methodology involves utilizing inter-frame data in addition to key-frame data, which effectively doubles the batch size. For a fair comparison, we increase the batch size of SLidR twofold. Additionally, to validate the effectiveness of the PPM's misalignment correction capability, we compare it with a baseline that minimizes misalignment, called the nearest alignment. The nearest alignment baseline utilizes unsynced data by employing both inter-frame LiDAR and inter-frame images but minimizes potential misalignment between inter-frame data by sampling inter-frame LiDAR and matching it with the closest timestamp's inter-frame image.
We find that our PPM module brings significant performance improvement (+3.7\%) compared to other baselines.
As shown in~\Fref{figure:supp_fig2}, we observed that the qualitative results improve as our treatment is applied sequentially.

\begin{table}[t]
    \caption{\textbf{Impact of data utilization and PPM.} To assess the effect of the proposed PPM module, we build two baseline methods, (A) and (B). Method (A) is the reproduced SLidR with the Cartesian coordinate. For a fair comparison, we have increased the batch size of SLidR, since (A) is trained on synced data only. Method (B) also utilizes the unsynced data and applies the nearest alignment to minimize misalignment between image and point cloud. With utilizing unsynced data, the proposed PPM module leads to significant performance improvement.}
    \centering
    \resizebox{0.8\linewidth}{!}{
    \begin{tabular}{c@{\hskip 0.2in}c@{\hskip 0.2in}c@{\hskip 0.2in}c@{\hskip 0.2in}c}
    \toprule
    \multirow{2}{*}{Methods} & \multirow{2}{*}{Data utilization} & \multirow{2}{*}{Alignment method} & nuScenes \\
    \cmidrule(lr){4-4}
    & & & Lin. Prob. (100\%) \\
    \midrule
    (A) & Synced & None  & 41.6 \\ 
    (B) & Synced + Unsynced & Nearest & 41.5 \\ 
    Ours & Synced + Unsynced & PPM & \textbf{45.2} \\ 
    \bottomrule
    \end{tabular}}
    \label{table:ablation_ppm}
\end{table}

% \begin{wraptable}{r}{5.5cm}
% \caption{A wrapped table going nicely inside the text.}\label{wrap-tab:1}
% \begin{tabular}{ccc}\\\toprule  
% Header-1 & Header-1 & Header-1 \\\midrule
% 2 &3 & 5\\  \midrule
% 2 &3 & 5\\  \midrule
% 2 &3 & 5\\  \bottomrule
% \end{tabular}
% \end{wraptable} 

% \begin{wraptable}{r}{5.5cm}

% \begin{table}[t]
%     \caption{\textbf{Impact of unsynced data and PPM.}}
%     \centering
%        \begin{tabular}{c@{\hskip 0.2in}c@{\hskip 0.2in}c@{\hskip 0.2in}c}
%         \toprule
%         \multirow{2}{*}{Data utilization} & Alignment method & nuScenes \\
%         \cmidrule(lr){3-3}
%         & & Lin. Prob. (100\%) \\
%         \midrule
%         Synced & None  & 41.6 \\ 
%         Synced + Unsynced & Nearest & 41.5 \\ 
%         Synced + Unsynced & PPM & \textbf{45.2} \\ 
%         \bottomrule
%     \end{tabular}
%     \label{table10}
% \end{table}
% We conduct ablation studies to validate the proposed PPM module designed to utilize the interframe data for self-supervised learning by 2D-to-3D. We fine-tune the pre-trained 3D backbone network on the nuScenes dataset with 1\% annotations and measure the mIoU metric on the 3D semantic segmentation task in nuScenes.
% For the comparison, we provide the reference experiments that employs the keyframe data only or the interframe data by using the 3D-to-2D projection $T$ only. With the proposed PPM module, using $T$ solely can not able to handle outliers caused by moving objects.
% As shown in Table~\ref{table6}, without the PPM module, na\"ively
% utilizing the inter-frame data shows marginal performance improvement compared to the experiment using the keyframe data only. In contrast, adopting the PPM module notably improves the fine-tuning performance. These results show that the PPM module is encouraged to be accompanied to leverage the inter-frame data effectively.

% We conduct ablation studies by measuring the mIoU on the validation set of nuScenes. Without the PPM module that handles moving points in the inter-frame data, performance improvement is marginal compared to the experiment using the keyframe data only. In contrast, utilizing the inter-frame data equipped with the PPM module demonstrates significant performance improvement.

\subsection{Transfer to Semantic Segmentation}\label{sec:segmentation}

We evaluate our pre-trained 3D backbone by fine-tuning it on a 3D semantic segmentation task. 
We choose the nuScenes~\cite{caesar2020nuscenes} dataset with 16 semantic categories and the SemanticKITTI~\cite{behley2019semantickitti} dataset with 19 semantic categories. 
We measure the Mean Intersection over Union (mIoU) metric for evaluation. 
Following common fine-tuning settings~\cite{sautier2022image,mahmoud2023self} we evaluate the performance on the mini-validation set of each dataset.  
For nuScenes dataset, we utilize two evaluation protocols: 1) linear probing with fully exploiting annotations and 2) fine-tuning with different portions of annotations, \ie $\{1\%, 5\%, 10\%, 25\%, 100\%\}$.
For the SemanticKITTI dataset, we report the fine-tuning results with few-shot
$\{1\%, 5\%, 10\%\}$ settings.

\begin{table*}[t]
\centering
    \caption{\textbf{3D semantic segmentation results on nuScenes and SemanticKITTI validation sets.} We compare our method with the existing 3D representation learning methods using the nuScenes and SemanticKITTI datasets. Our method surpasses the existing methods across all metrics. The table is horizontally partitioned based on whether (Top) fully-unsupervised methods or (Bottom) SAM were used.
    % Additionally, our method is superior in all but one metric compared to Seal, which utilizes Vision Foundation Models (VFM) with a superpixel segmenter. Our method, Ours-SAM, which incorporates VFM, shows the same trend.
    % Fine-tuning performance depending on 
    }
    \resizebox{1\linewidth}{!}{
    \begin{tabular}{m{3cm}ccccccccc}
    \toprule
    \multirow{2}{*}{Method} & \multicolumn{6}{c}{nuScenes} & \multicolumn{3}{c}{SemanticKITTI} \\
    \cmidrule(lr){2-7} \cmidrule(lr){8-10}
    & Lin. Prob. & 1\% & 5\% & 10\% & 25\% & 100\% & 1\% & 5\% & 10\% \\
    \midrule
    Random & 8.1 & 30.3 & 47.7 & 56.6 & 64.8 & 74.2 & 39.5 & 52.1 & 55.6 \\ 
    PointContrast~\cite{xie2020pointcontrast} & 21.9 & 32.5~\textcolor{ballblue}{(+2.2)}  & - & 57.1~\textcolor{ballblue}{(+0.5)} & - & 74.3~\textcolor{ballblue}{(+0.1)} & 41.1~\textcolor{ballblue}{(+1.6)} & - & - \\ 
    DepthContrast~\cite{zhang2021self} & 22.1 & 31.7~\textcolor{ballblue}{(+1.4)}  & - & 57.3~\textcolor{ballblue}{(+0.7)} & - & 74.1~\textcolor{red}{(-0.1)} & 41.5~\textcolor{ballblue}{(+2.0)} & - & - \\ 
    % TARL~\cite{nunes2023temporal} & 34.6 & - & 54.1 & - & - & 44.6 & - & 54.1 \\ 
    PPKT~\cite{liu2021learning} & 36.4 & 37.8~\textcolor{ballblue}{(+7.5)} & 51.7\textcolor{ballblue}{(+4.0)} & 59.2~\textcolor{ballblue}{(+2.6)} & 66.8~\textcolor{ballblue}{(+2.0)} & 73.8~\textcolor{red}{(-0.4)} & 43.9~\textcolor{ballblue}{(+4.4)} & 53.1~\textcolor{ballblue}{(+1.0)} & 57.3~\textcolor{ballblue}{(+1.7)} \\ 
    SLidR~\cite{sautier2022image} & 38.8 & 38.2~\textcolor{ballblue}{(+7.9)} & 52.2~\textcolor{ballblue}{(+4.5)} & 58.8~\textcolor{ballblue}{(+2.2)} & 66.2~\textcolor{ballblue}{(+1.4)} & 74.6~\textcolor{ballblue}{(+0.4)} & 44.6~\textcolor{ballblue}{(+5.1)} & 52.6~\textcolor{ballblue}{(+0.5)} & 56.0~\textcolor{ballblue}{(+0.4)} \\ 
    ST-SLidR~\cite{mahmoud2023self} & 40.4 & 40.7~\textcolor{ballblue}{(+10.4)} & 54.6~\textcolor{ballblue}{(+6.9)} & 60.7~\textcolor{ballblue}{(+4.1)} & 67.7~\textcolor{ballblue}{(+2.9)} & 75.1~\textcolor{ballblue}{(+0.9)} & 44.7~\textcolor{ballblue}{(+5.2)} & - & - \\ 
    TriCC~\cite{pang2023unsupervised}  & 38.0 & 41.2~\textcolor{ballblue}{(+10.9)} & 54.1~\textcolor{ballblue}{(+6.4)} & 60.4~\textcolor{ballblue}{(+3.8)} & 67.6~\textcolor{ballblue}{(+2.8)} & 75.6~\textcolor{ballblue}{(+1.4)} & 45.9~\textcolor{ballblue}{(+6.4)} & 55.9~\textcolor{ballblue}{(+3.8)} & 59.0 ~\textcolor{ballblue}{(+3.4)}\\ 
    \textbf{Ours} & 
    \textbf{45.2} &
    \textbf{42.7}~\textbf{\textcolor{ballblue}{(+12.4)}} &
    \textbf{56.8}~\textbf{\textcolor{ballblue}{(+9.1)}} &
    \textbf{63.3}~\textbf{\textcolor{ballblue}{(+6.7)}} &
    \textbf{69.8}~\textbf{\textcolor{ballblue}{(+5.0)}} &
    \textbf{75.7}~\textbf{\textcolor{ballblue}{(+1.5)}} & 
    \textbf{50.3}~\textbf{\textcolor{ballblue}{(+10.8)}} &
    \textbf{61.1}~\textbf{\textcolor{ballblue}{(+9.0)}} & 
    \textbf{63.3}~\textbf{\textcolor{ballblue}{(+7.7)}} \\ 
    \midrule
    \midrule
    Seal~\cite{liu2023segment} & 44.9 & ~\textbf{45.8}~\textbf{\textcolor{ballblue}{(+15.5)}} & 55.6~\textcolor{ballblue}{(+7.9)} & 62.9~\textcolor{ballblue}{(+6.3)} & 68.4~\textcolor{ballblue}{(+3.6)}& 75.6~\textcolor{ballblue}{(+1.4)} & 46.6~\textcolor{ballblue}{(+7.1)} & - & - \\ 
    \textbf{Ours-SAM} &
    \textbf{48.0} &
    43.0~\textcolor{ballblue}{(+12.7)} &
    \textbf{57.1}~\textbf{\textcolor{ballblue}{(+9.4)}} &
    \textbf{64.7}~\textbf{\textcolor{ballblue}{(+8.1)}} &
    \textbf{70.0}~\textbf{\textcolor{ballblue}{(+5.2)}} &
    \textbf{75.9}~\textbf{\textcolor{ballblue}{(+1.7)}} &
    \textbf{52.1}~\textbf{\textcolor{ballblue}{(+12.6)}} &
    \textbf{61.5}~\textbf{\textcolor{ballblue}{(+9.4)}} &
    \textbf{63.6}~\textbf{\textcolor{ballblue}{(+8.0)}} \\ 
    \bottomrule
    \end{tabular}}
    \label{table:segmentation}
\end{table*}
\vspace{3mm} %segmentation
\addtolength{\tabcolsep}{2pt}    
\begin{table}[t]
    \caption{\textbf{Few-shot 3D object detection results on KITTI.} Using the KITTI dataset, we fine-tune our model with few labeled dataset and evaluate the mAP@R40 metric. Our method outperforms the recent Image-to-LiDAR distillation methods, \ie, PPKT~\cite{liu2021learning}, SLidR~\cite{sautier2022image}, and TriCC~\cite{pang2023unsupervised} by a large margin and achieves on average more than a two-fold performance increase. \dag\ denotes KITTI pretraining, \ddag\ denotes Waymo~\cite{sun2020scalability} pretraining.
    The table is horizontally partitioned by different epoch settings.
    }
\centering
    \resizebox{1.0\linewidth}{!}{
    \begin{tabular}{l c ccc ccc ccc}
    \toprule
    \multirow{2}{*}{Method} & \multirow{2}{*}{Epoch} &
    \multicolumn{3}{c}{Fine-tune (5\%)} & \multicolumn{3}{c}{Fine-tune (10\%)} & \multicolumn{3}{c}{Fine-tune (20\%)} \\
    \cmidrule(lr){3-5}
    \cmidrule(lr){6-8}
    \cmidrule(lr){9-11}
    & & Easy & Moderate & Hard & Easy & Moderate & Hard & Easy & Moderate & Hard \\
    \midrule
    Random & 50 & 73.7 & 56.6 & 50.7 & 74.6 & 58.8 & 53.9 & 77.9 & 63.7 & 59.2 \\ 
    PPKT~\cite{liu2021learning} & 50 & 75.7~\textcolor{ballblue}{(+2.0)} & 59.6~\textcolor{ballblue}{(+3.0)} & 54.4~\textcolor{ballblue}{(+3.7)} & 78.3~\textcolor{ballblue}{(+3.7)} & 63.7~\textcolor{ballblue}{(+4.9)} & 58.4~\textcolor{ballblue}{(+4.5)} & 78.9~\textcolor{ballblue}{(+1.0)} & 64.8~\textcolor{ballblue}{(+1.1)} & 59.9~\textcolor{ballblue}{(+0.7)} \\ 
    SLidR~\cite{sautier2022image} & 50 & 74.5~\textcolor{ballblue}{(+0.8)} & 58.8~\textcolor{ballblue}{(+2.2)} & 52.9~\textcolor{ballblue}{(+2.2)} & 78.1~\textcolor{ballblue}{(+3.5)} & 63.5~\textcolor{ballblue}{(+4.7)} & 58.3~\textcolor{ballblue}{(+4.4)} & 77.6~\textcolor{red}{(-0.3)} & 63.8~\textcolor{ballblue}{(+0.1)} & 59.2 \\ 
    \textbf{Ours} & 50 & \textbf{84.3}~\textcolor{ballblue}{(+10.6)} & \textbf{69.3}~\textcolor{ballblue}{(+12.7)} & \textbf{64.0}~\textcolor{ballblue}{(+13.3)} & \textbf{84.0}~\textcolor{ballblue}{(+9.4)} & \textbf{70.9}~\textcolor{ballblue}{(+12.1)} & \textbf{66.1}~\textcolor{ballblue}{(+12.2)} & \textbf{84.0}~\textcolor{ballblue}{(+6.1)} & \textbf{70.9}~\textcolor{ballblue}{(+7.2)} & \textbf{66.1}~\textcolor{ballblue}{(+6.9)} \\
    \midrule
    TriCC~\cite{pang2023unsupervised} & 20 & 77.9~\textcolor{ballblue}{(+4.2)} & 61.3~\textcolor{ballblue}{(+4.7)} & 56.2~\textcolor{ballblue}{(+5.5)} & 79.6~\textcolor{ballblue}{(+5.0)} & 64.6~\textcolor{ballblue}{(+5.8)} & 59.3~\textcolor{ballblue}{(+5.4)} & 80.0~\textcolor{ballblue}{(+2.1)} & 65.9~\textcolor{ballblue}{(+2.2)} & 60.7~\textcolor{ballblue}{(+1.5)} \\
    GPC\textsuperscript{\dag}~\cite{pan2023gpc} & 80 & - & 62.7~\textcolor{ballblue}{(+6.1)} & - & - & 66.4~\textcolor{ballblue}{(+7.6)} & - & - & 70.7~\textcolor{ballblue}{(+7.0)} & - \\
    GPC\textsuperscript{\ddag}~\cite{pan2023gpc} & 12 & - & 63.9~\textcolor{ballblue}{(+7.3)} & - & - & 67.0~\textcolor{ballblue}{(+8.2)} & - & - & 70.1~\textcolor{ballblue}{(+6.4)} & - \\
    \textbf{Ours} & 20 & 83.4~\textcolor{ballblue}{(+9.7)} & \textbf{69.3}~\textcolor{ballblue}{(+12.7)} & 63.7~\textcolor{ballblue}{(+13.0)} & 82.5~\textcolor{ballblue}{(+7.9)} & 70.0~\textcolor{ballblue}{(+11.2)} & 65.5~\textcolor{ballblue}{(+11.6)} & 82.5~\textcolor{ballblue}{(+4.6)} & 70.0~\textcolor{ballblue}{(+6.3)} & 65.5~\textcolor{ballblue}{(+6.3)} \\
    \bottomrule
    \end{tabular}}
    \label{table:detection}
\end{table}
\addtolength{\tabcolsep}{1pt}    

% \begin{table}[t]
% \centering
%     \resizebox{0.70\linewidth}{!}{
%     \begin{tabular}{c@{\hskip 0.2in}c@{\hskip 0.2in}c@{\hskip 0.2in}c}
%     \toprule
%     \multirow{2}{*}{Methods} & \multirow{2}{*}{Coordinate} & \multirow{2}{*}{Voxel size ($cm$)} & nuScenes \\
%     \cmidrule(lr){4-4}
%     & & & Lin Prob (100\%) \\
%     \midrule
%     SLidR~\cite{sautier2022image} & Cylindrical & 10 & 38.8 \\
%     (A) & Cartesian & 10 & 40.8 \\
%     (B) & Cartesian & 5 & 40.9 \\
%     \bottomrule
%     \end{tabular}}
%     \caption{\textbf{Ablation study on the LiDAR coordinate system and voxel size.} 
%  }
%     \label{table10}
% \end{table} %detection
\begin{table*}[t]
    \caption{\textbf{Per-class fine-tuning performance on nuScenes with 1\% of annotation.} We evaluate the per-class IoU on nuScenes. Compared to competing methods, our method achieves the best performance in the average class. Specifically, it significantly boosts the performance for classes that are challenging to learn due to their smaller resolution, \eg, bicycle, pedestrian, traffic cone.
    }
\centering
    \resizebox{1\linewidth}{!}{
    \begin{tabular}{ccccccccccccccccccc}
    \toprule
    {Method} & \rotatebox{60}{barrier} & \rotatebox{60}{bicycle} & \rotatebox{60}{bus} & \rotatebox{60}{car} & \rotatebox{60}{const. veh.} & \rotatebox{60}{motorcycle} & \rotatebox{60}{pedestrian} & \rotatebox{60}{traffic cone} & \rotatebox{60}{trailer} & \rotatebox{60}{truck} & \rotatebox{60}{driv. surf.} & \rotatebox{60}{other flat} & \rotatebox{60}{sidewalk} & \rotatebox{60}{terrian} & \rotatebox{60}{manmade} & \rotatebox{60}{vegetation} & \rotatebox{60}{\textbf{mIoU}} \\
    \midrule
    Random & 0.0 & 0.0 & 8.1 & 65.0 & 0.1 & 6.6 & 21.0 & 9.0 & 9.3 & 25.8 & 89.5 & 14.8 & 41.7 & 48.7 & 72.4 & 73.3 & 30.3 \\ 
    PointContrast~\cite{xie2020pointcontrast} & 0.0  & 1.0 & 5.6 & 67.4 & 0.0 & 3.3 & 31.6 & 5.6 & 12.1 & 30.8 & 91.7 & 21.9 & 48.4 & 50.8 & 75.0 & 74.6 & 32.5 \\ 
    DepthContrast~\cite{zhang2021self} & 0.0 & 0.6 & 6.5 & 64.7 & 0.2 & 5.1 & 29.0 & 9.5 & 12.1 & 29.9 & 90.3 & 17.8 & 44.4 & 49.5 & 73.5 & 74.0 & 31.7 \\ 
    % TARL~\cite{nunes2023temporal} & 0.0 & 0.4 & 8.2 & 72.2 & 0.3 & 13.3 & 40.3 & 11.2 & 9.1 & 35.4 & 89.3 & 22.5 & 42.2 & 57.1 & 76.1 & 78.2 & 34.7 \\ 
    PPKT~\cite{liu2021learning} & 0.0 & 2.2 & 20.7 & 75.4 & 1.2 & 13.2 & 45.6 & 8.5 & 17.5 & 38.4 & 92.5 & 19.2 & 52.3 & 56.8 & 80.1 & 80.9 & 37.8 \\ 
    SLidR~\cite{sautier2022image} & 0.0 & 3.1 & 15.2 & 72.0 & 0.9 & 18.8 & 43.2 & 12.5 & 14.7 & 33.3 & 92.8 & 29.4 & 54.0 & 61.0 & 80.2 & 81.9 & 38.3 \\ 
    ST-SLidR~\cite{mahmoud2023self} & 0.0 & 2.7 & 16.0 & 74.5 & \textbf{3.2} & \textbf{25.4} & 50.9 & 20.0 & \textbf{17.7} & 40.2 & 92.0 & 30.7 & 54.2 & 61.1 & 80.5 & \textbf{82.9} & 40.7 \\ 
    TriCC~\cite{pang2023unsupervised}  & 0.0 & 2.6 & 20.7 & 73.6 & 0.3 & 18.9 & 49.2 & 22.0 & 16.9 & 33.4 & \textbf{94.5} & \textbf{43.1} & \textbf{57.2} & \textbf{62.1} & \textbf{82.3} & 82.6 & 41.2 \\ 
    \midrule
    Ours & 0.0 & \textbf{3.4} & \textbf{26.0} & \textbf{77.4} & 2.4 & 21.4 & \textbf{61.4} & \textbf{30.9} & 17.5 & \textbf{44.4} & 92.2 & 28.3 & 54.1 & \textbf{62.1} & 79.6 & 81.6 & \textbf{42.7} \\ 
    \bottomrule
    \end{tabular}}
    \label{table8}

\end{table*}

In~\Tref{table:segmentation}, we evaluate our learned representations on the 3D semantic segmentation task.
To demonstrate, we establish a Random model by randomly initializing it without using the pre-trained weight from the 3D network.
For the nuScenes dataset~\cite{caesar2020nuscenes},
we find that all the self-supervised models initialized with pre-trained weights show improvements over the model initialized with random weight.
Interestingly, we observe that our method outperforms all the competing methods by a large margin, \eg, +37.1\% for linear probing and +12.4\% for fine-tuning with 1\% annotations. 
We also evaluate the fine-tuning performance on the SemanticKITTI dataset~\cite{behley2019semantickitti} to verify the effectiveness.
Although our model is only pre-trained on the nuScenes dataset, 
our model surpasses all the competing methods across varying annotation settings.
This demonstrates that our method can generalize to unseen datasets and
different volumes of annotations.
These results reveal the dissonance of the existing baselines' design choices, and our simple treatments of input interface and data utilization effectively resolve the issues.
Additionally, we compare our methodology with Seal~\cite{liu2023segment}, which utilizes Vision Foundation Models (VFM) with a superpixel segmenter instead of SLIC~\cite{achanta2012slic}. 
Despite using SLIC as is, our method shows superior performance in all metrics except one compared to Seal. Furthermore, when we replace SLIC with one of the VFMs, Segment Anything Models (SAM)~\cite{kirillov2023segment}, in our methodology, dubbed Ours-SAM, we observe a consistent overall improvement in performance. This suggests that our methodology, proposing a way to utilize raw data itself, can be applied to other methods to achieve consistently improved performance.

\subsection{Transfer to Object Detection}\label{sec:detection}
To validate the usefulness of the 3D representation learned by the proposed method, we report performance on another challenging downstream task, few-shot 3D object detection. For the few-shot 3D object detection dataset, we adopt the KITTI dataset~\cite{Geiger2012CVPR}. 
For a fair comparison, we exactly follow the fine-tuning implementation details of SLidR~\cite{sautier2022image}. Table~\ref{table:detection} shows the mAP@R40 performance of networks fine-tuned on 5\%, 10\%, and 20\% annotations on the difficulty levels defined by the KITTI dataset, \eg, easy, moderate, and hard. For a fair comparison with TriCC~\cite{pang2023unsupervised}, we provide the performance of the pre-trained network for 20 epochs as well.
Our method archives the best performance across all metrics and surpasses the other Image-to-LiDAR distillation methods, \eg, PPKT~\cite{liu2021learning}, SLiDR~\cite{sautier2022image}, TriCC~\cite{pang2023unsupervised} by a large margin regardless of the number of epochs. Compared to the improvements achieved by previous works at random, we achieve, on average, more than a two-fold performance increase with simple treatments, and in the fine-tune 5\% moderate setting, we achieve a four-fold performance improvement. 
Our method also shows a significant performance improvement compared to GPC~\cite{pan2023gpc} pre-trained on KITTI and Waymo, which have similar or identical LiDAR characteristics to downstream dataset \ie KITTI.

% \subsection{Effects of the Volume of Annotations}\label{sec:volume}
% Because of the virtue of data utility, our method achieves the state-of-art performance in the fine-tuning and linear probing on the nuScene and SemanticKITTI datasets. However, a question can arise regarding the effectiveness of our method as the volume of annotation increases in fine-tuning. To answer this question, we conduct fine-tuning experiments on the nuScene and SemanticKITTI datasets by varying the volume of annotation. As shown in  Table~\ref{table3}, we observe that our method shows the best performance on the nuScenes and SemanticKITTI datasets regardless of the volume of annotations. These results validate the efficacy of our proposed method in enhancing data utility, demonstrating its effectiveness not only with a limited amount of annotation but also with full annotations.

% \input{tables/table8}

\subsection{Per-class IoU}
We report the 3D semantic segmentation performance of per-class by fine-tuning 1\% on the nuScenes dataset (see Table~\ref{table8}).
Compared to the competing methods, Ours achieves the best performance at the average of classes and is ranked first in 16 classes.
By balancing the overall quantization error to increase data resolution, performance across all classes improves. Notably, this leads to a significant enhancement in performance for classes that are difficult to learn due to their small resolution, such as bicycle, pedestrian, and traffic cone. Especially with treatments focused on data utilization, we show good performance compared to ST-SLidR~\cite{mahmoud2023self}, which is designed with a loss aware of rare classes (e.g., bicycle, motorcycle, pedestrian, and traffic cone) that have a small number of points.

\section{Conclusion}
We delve into 3D representation learning, especially in image-to-LiDAR distillation, and find that the prior arts overlook the fundamental designs, \eg, the LiDAR coordinate system, quantization, and data under-utilization.
Compared to the existing works focusing on their own design of loss functions,
we propose simple remedies for spatial and temporal designs.
Spatially, we change the domain of quantization from cylindrical to Cartesian, which ensures uniform quantization error regardless of distance and prevents the degradation of spatial resolution.
Temporally, we propose a PPM module that enables the model to utilize the unsynced data by aligning mismatched pixel-point paired data.
Our spatial and temporal treatments facilitate the synergies and exhibit 
significant performance improvements, \ie, state-of-the-art performance under the same benchmark setting in two downstream tasks: 3D semantic segmentation and 3D object detection.
We hope that our work provides common ground by offering our new baseline and input protocol to contribute to future image-to-LiDAR distillation research and help pave the way for further development.

\subsubsection{\ackname}
This project was supported by RideFlux and also supported by the Institute of Information \& communications Technology Planning \& Evaluation (IITP) grant funded by the Korea government(MSIT) (No.RS-2022-II220124, Development of Artificial Intelligence Technology for Self-Improving CompetencyAware Learning Capabilities; No. 2020-0-00004, Development of Previsional Intelligence based on Long-term Visual Memory Network; No.RS-2020-II201336, Artificial Intelligence Graduate School Program(UNIST))

% This work was supported by Institute of Information \& communications Technology Planning \& Evaluation (IITP) grant funded by the Korea government (MSIT) (No.RS-2020-II201336, Artificial Intelligence Graduate School Program(UNIST)).

%c
\bibliographystyle{splncs04}
\bibliography{main}

\begin{thebibliography}{10}
\providecommand{\url}[1]{\texttt{#1}}
\providecommand{\urlprefix}{URL }
\providecommand{\doi}[1]{https://doi.org/#1}

\bibitem{achanta2012slic}
Achanta, R., Shaji, A., Smith, K., Lucchi, A., Fua, P., S{\"u}sstrunk, S.: Slic superpixels compared to state-of-the-art superpixel methods. IEEE Transactions on Pattern Analysis and Machine Intelligence (TPAMI)  \textbf{34}(11),  2274--2282 (2012)

\bibitem{bardes2021vicreg}
Bardes, A., Ponce, J., LeCun, Y.: Vicreg: Variance-invariance-covariance regularization for self-supervised learning. arXiv preprint arXiv:2105.04906  (2021)

\bibitem{behley2019semantickitti}
Behley, J., Garbade, M., Milioto, A., Quenzel, J., Behnke, S., Stachniss, C., Gall, J.: Semantickitti: A dataset for semantic scene understanding of lidar sequences. In: IEEE International Conference on Computer Vision (ICCV). pp. 9297--9307 (2019)

\bibitem{caesar2020nuscenes}
Caesar, H., Bankiti, V., Lang, A.H., Vora, S., Liong, V.E., Xu, Q., Krishnan, A., Pan, Y., Baldan, G., Beijbom, O.: nuscenes: A multimodal dataset for autonomous driving. In: IEEE Conference on Computer Vision and Pattern Recognition (CVPR). pp. 11621--11631 (2020)

\bibitem{campello2013density}
Campello, R.J., Moulavi, D., Sander, J.: Density-based clustering based on hierarchical density estimates. In: Pacific-Asia conference on knowledge discovery and data mining. pp. 160--172. Springer (2013)

\bibitem{chen2020simple}
Chen, T., Kornblith, S., Norouzi, M., Hinton, G.: A simple framework for contrastive learning of visual representations. In: International Conference on Machine Learning (ICML). pp. 1597--1607. PMLR (2020)

\bibitem{chen2020improved}
Chen, X., Fan, H., Girshick, R., He, K.: Improved baselines with momentum contrastive learning. arXiv preprint arXiv:2003.04297  (2020)

\bibitem{chen2021shape}
Chen, Y., Liu, J., Ni, B., Wang, H., Yang, J., Liu, N., Li, T., Tian, Q.: Shape self-correction for unsupervised point cloud understanding. In: Proceedings of the IEEE/CVF International Conference on Computer Vision. pp. 8382--8391 (2021)

\bibitem{cheng20212}
Cheng, R., Razani, R., Taghavi, E., Li, E., Liu, B.: 2-s3net: Attentive feature fusion with adaptive feature selection for sparse semantic segmentation network. In: Proceedings of the IEEE/CVF conference on computer vision and pattern recognition. pp. 12547--12556 (2021)

\bibitem{choy20194d}
Choy, C., Gwak, J., Savarese, S.: 4d spatio-temporal convnets: Minkowski convolutional neural networks. In: IEEE Conference on Computer Vision and Pattern Recognition (CVPR). pp. 3075--3084 (2019)

\bibitem{Geiger2012CVPR}
Geiger, A., Lenz, P., Urtasun, R.: Are we ready for autonomous driving? the kitti vision benchmark suite. In: Conference on Computer Vision and Pattern Recognition (CVPR) (2012)

\bibitem{geiger2012we}
Geiger, A., Lenz, P., Urtasun, R.: Are we ready for autonomous driving? the kitti vision benchmark suite. In: IEEE Conference on Computer Vision and Pattern Recognition (CVPR). pp. 3354--3361. IEEE (2012)

\bibitem{gojcic2021weakly}
Gojcic, Z., Litany, O., Wieser, A., Guibas, L.J., Birdal, T.: Weakly supervised learning of rigid 3d scene flow. In: Proceedings of the IEEE/CVF conference on computer vision and pattern recognition. pp. 5692--5703 (2021)

\bibitem{hadsell2006dimensionality}
Hadsell, R., Chopra, S., LeCun, Y.: Dimensionality reduction by learning an invariant mapping. In: 2006 IEEE computer society conference on computer vision and pattern recognition (CVPR'06). vol.~2, pp. 1735--1742. IEEE (2006)

\bibitem{he2020momentum}
He, K., Fan, H., Wu, Y., Xie, S., Girshick, R.: Momentum contrast for unsupervised visual representation learning. In: Proceedings of the IEEE/CVF conference on computer vision and pattern recognition. pp. 9729--9738 (2020)

\bibitem{he2016deep}
He, K., Zhang, X., Ren, S., Sun, J.: Deep residual learning for image recognition. In: Proceedings of the IEEE conference on computer vision and pattern recognition. pp. 770--778 (2016)

\bibitem{henaff2020data}
Henaff, O.: Data-efficient image recognition with contrastive predictive coding. In: International conference on machine learning. pp. 4182--4192. PMLR (2020)

\bibitem{hou2021exploring}
Hou, J., Graham, B., Nie{\ss}ner, M., Xie, S.: Exploring data-efficient 3d scene understanding with contrastive scene contexts. In: Proceedings of the IEEE/CVF Conference on Computer Vision and Pattern Recognition. pp. 15587--15597 (2021)

\bibitem{huang2022dynamic}
Huang, S., Gojcic, Z., Huang, J., Wieser, A., Schindler, K.: Dynamic 3d scene analysis by point cloud accumulation. In: European Conference on Computer Vision. pp. 674--690. Springer (2022)

\bibitem{huang2021spatio}
Huang, S., Xie, Y., Zhu, S.C., Zhu, Y.: Spatio-temporal self-supervised representation learning for 3d point clouds. In: Proceedings of the IEEE/CVF International Conference on Computer Vision. pp. 6535--6545 (2021)

\bibitem{kirillov2023segment}
Kirillov, A., Mintun, E., Ravi, N., Mao, H., Rolland, C., Gustafson, L., Xiao, T., Whitehead, S., Berg, A.C., Lo, W.Y., et~al.: Segment anything. arXiv preprint arXiv:2304.02643  (2023)

\bibitem{lee2022patchwork++}
Lee, S., Lim, H., Myung, H.: Patchwork++: Fast and robust ground segmentation solving partial under-segmentation using 3d point cloud. In: 2022 IEEE/RSJ International Conference on Intelligent Robots and Systems (IROS). pp. 13276--13283. IEEE (2022)

\bibitem{liu2023segment}
Liu, Y., Kong, L., Cen, J., Chen, R., Zhang, W., Pan, L., Chen, K., Liu, Z.: Segment any point cloud sequences by distilling vision foundation models. arXiv preprint arXiv:2306.09347  (2023)

\bibitem{liu2021learning}
Liu, Y.C., Huang, Y.K., Chiang, H.Y., Su, H.T., Liu, Z.Y., Chen, C.T., Tseng, C.Y., Hsu, W.H.: Learning from 2d: Contrastive pixel-to-point knowledge transfer for 3d pretraining. arXiv preprint arXiv:2104.04687  (2021)

\bibitem{mahmoud2023self}
Mahmoud, A., Hu, J.S., Kuai, T., Harakeh, A., Paull, L., Waslander, S.L.: Self-supervised image-to-point distillation via semantically tolerant contrastive loss. In: IEEE Conference on Computer Vision and Pattern Recognition (CVPR). pp. 7102--7110 (2023)

\bibitem{misra2020self}
Misra, I., Maaten, L.v.d.: Self-supervised learning of pretext-invariant representations. In: Proceedings of the IEEE/CVF conference on computer vision and pattern recognition. pp. 6707--6717 (2020)

\bibitem{nunes2022segcontrast}
Nunes, L., Marcuzzi, R., Chen, X., Behley, J., Stachniss, C.: Segcontrast: 3d point cloud feature representation learning through self-supervised segment discrimination. IEEE Robotics and Automation Letters  \textbf{7}(2),  2116--2123 (2022)

\bibitem{nunes2023temporal}
Nunes, L., Wiesmann, L., Marcuzzi, R., Chen, X., Behley, J., Stachniss, C.: Temporal consistent 3d lidar representation learning for semantic perception in autonomous driving. In: IEEE Conference on Computer Vision and Pattern Recognition (CVPR). pp. 5217--5228 (2023)

\bibitem{oord2018representation}
Oord, A.v.d., Li, Y., Vinyals, O.: Representation learning with contrastive predictive coding. arXiv preprint arXiv:1807.03748  (2018)

\bibitem{pan2023gpc}
Pan, T.Y., Ma, C., Chen, T., Phoo, C.P., Luo, K.Z., You, Y., Campbell, M., Weinberger, K.Q., Hariharan, B., Chao, W.L.: Pre-training lidar-based 3d object detectors through colorization (2024)

\bibitem{pang2023unsupervised}
Pang, B., Xia, H., Lu, C.: Unsupervised 3d point cloud representation learning by triangle constrained contrast for autonomous driving. In: IEEE Conference on Computer Vision and Pattern Recognition (CVPR). pp. 5229--5239 (2023)

\bibitem{puy2024three}
Puy, G., Gidaris, S., Boulch, A., Sim{\'e}oni, O., Sautier, C., P{\'e}rez, P., Bursuc, A., Marlet, R.: Three pillars improving vision foundation model distillation for lidar. In: Proceedings of the IEEE/CVF Conference on Computer Vision and Pattern Recognition. pp. 21519--21529 (2024)

\bibitem{rusinkiewicz2001efficient}
Rusinkiewicz, S., Levoy, M.: Efficient variants of the icp algorithm. In: Proceedings third international conference on 3-D digital imaging and modeling. pp. 145--152. IEEE (2001)

\bibitem{sauder2019self}
Sauder, J., Sievers, B.: Self-supervised deep learning on point clouds by reconstructing space. Advances in Neural Information Processing Systems  \textbf{32} (2019)

\bibitem{sautier2022image}
Sautier, C., Puy, G., Gidaris, S., Boulch, A., Bursuc, A., Marlet, R.: Image-to-lidar self-supervised distillation for autonomous driving data. In: IEEE Conference on Computer Vision and Pattern Recognition (CVPR). pp. 9891--9901 (2022)

\bibitem{sun2020scalability}
Sun, P., Kretzschmar, H., Dotiwalla, X., Chouard, A., Patnaik, V., Tsui, P., Guo, J., Zhou, Y., Chai, Y., Caine, B., et~al.: Scalability in perception for autonomous driving: Waymo open dataset. In: IEEE Conference on Computer Vision and Pattern Recognition (CVPR). pp. 2446--2454 (2020)

\bibitem{tang2020searching}
Tang, H., Liu, Z., Zhao, S., Lin, Y., Lin, J., Wang, H., Han, S.: Searching efficient 3d architectures with sparse point-voxel convolution. In: European Conference on Computer Vision (ECCV). pp. 685--702. Springer (2020)

\bibitem{tian2019contrastive}
Tian, Y., Krishnan, D., Isola, P.: Contrastive representation distillation. arXiv preprint arXiv:1910.10699  (2019)

\bibitem{wang2021unsupervised}
Wang, H., Liu, Q., Yue, X., Lasenby, J., Kusner, M.J.: Unsupervised point cloud pre-training via occlusion completion. In: Proceedings of the IEEE/CVF international conference on computer vision. pp. 9782--9792 (2021)

\bibitem{wu2023spatiotemporal}
Wu, Y., Zhang, T., Ke, W., S{\"u}sstrunk, S., Salzmann, M.: Spatiotemporal self-supervised learning for point clouds in the wild. In: IEEE Conference on Computer Vision and Pattern Recognition (CVPR). pp. 5251--5260 (2023)

\bibitem{wu2018unsupervised}
Wu, Z., Xiong, Y., Yu, S.X., Lin, D.: Unsupervised feature learning via non-parametric instance discrimination. In: Proceedings of the IEEE conference on computer vision and pattern recognition. pp. 3733--3742 (2018)

\bibitem{xie2020pointcontrast}
Xie, S., Gu, J., Guo, D., Qi, C.R., Guibas, L., Litany, O.: Pointcontrast: Unsupervised pre-training for 3d point cloud understanding. In: European Conference on Computer Vision (ECCV). pp. 574--591. Springer (2020)

\bibitem{yin2022proposalcontrast}
Yin, J., Zhou, D., Zhang, L., Fang, J., Xu, C.Z., Shen, J., Wang, W.: Proposalcontrast: Unsupervised pre-training for lidar-based 3d object detection. In: European Conference on Computer Vision. pp. 17--33. Springer (2022)

\bibitem{zbontar2021barlow}
Zbontar, J., Jing, L., Misra, I., LeCun, Y., Deny, S.: Barlow twins: Self-supervised learning via redundancy reduction. In: International Conference on Machine Learning. pp. 12310--12320. PMLR (2021)

\bibitem{zhang2023simple}
Zhang, H., Li, F., Zou, X., Liu, S., Li, C., Yang, J., Zhang, L.: A simple framework for open-vocabulary segmentation and detection. In: Proceedings of the IEEE/CVF International Conference on Computer Vision. pp. 1020--1031 (2023)

\bibitem{zhang2021self}
Zhang, Z., Girdhar, R., Joulin, A., Misra, I.: Self-supervised pretraining of 3d features on any point-cloud. In: IEEE International Conference on Computer Vision (ICCV). pp. 10252--10263 (2021)

\bibitem{zhu2021cylindrical}
Zhu, X., Zhou, H., Wang, T., Hong, F., Ma, Y., Li, W., Li, H., Lin, D.: Cylindrical and asymmetrical 3d convolution networks for lidar segmentation. In: IEEE Conference on Computer Vision and Pattern Recognition (CVPR). pp. 9939--9948 (2021)

\bibitem{zou2023generalized}
Zou, X., Dou, Z.Y., Yang, J., Gan, Z., Li, L., Li, C., Dai, X., Behl, H., Wang, J., Yuan, L., et~al.: Generalized decoding for pixel, image, and language. In: Proceedings of the IEEE/CVF Conference on Computer Vision and Pattern Recognition. pp. 15116--15127 (2023)

\bibitem{zou2024segment}
Zou, X., Yang, J., Zhang, H., Li, F., Li, L., Wang, J., Wang, L., Gao, J., Lee, Y.J.: Segment everything everywhere all at once. Advances in Neural Information Processing Systems  \textbf{36} (2024)

\end{thebibliography}

% ---- supplementary material ----

%
\title{The Devil is in the Details: Simple Remedies for 
Image-to-LiDAR Representation Learning \\
\textmd{--- Supplementary Material ---}}
\titlerunning{Simple Remedies for Image-to-LiDAR Representation Learning}
\author{Wonjun Jo\inst{1}\orcidlink{0000-0003-3894-5483} \and
Kwon Byung-Ki\inst{2}\orcidlink{0000-0003-4187-7944} \and
Kim Ji-Yeon\inst{3}\orcidlink{0000-0002-7535-380X} \and
Hawook Jeong\inst{4}\orcidlink{0009-0003-5935-7055} \and
Kyungdon Joo\inst{5}\orcidlink{0000-0002-3920-9608} \and
Tae-Hyun Oh\inst{1,2,6}\orcidlink{0000-0003-0468-1571}
}
\authorrunning{W. Jo et al.}
\institute{Department of Electrical Engineering, POSTECH, South Korea \and Graduate School of AI, POSTECH, South Korea \and
Department of Convergence IT Engineering, POSTECH, South Korea \and
RideFlux Inc., South Korea \and
Artificial Intelligence Graduate School, UNIST, South Korea \and
Institute for Convergence Research and Education in Advanced Technology, Yonsei University, South Korea \\
\email{\{jo1jun,byungki.kwon,jiyeon.kim,taehyun\}@postech.ac.kr}, \email{hawook@rideflux.com}, \email{kyungdon@unist.ac.kr}
}
\maketitle              % typeset the header of the contribution

In this supplementary material, we provide the additional qualitative results (Sec.~\ref{sec:supp_qual}), 
additional experiments (Sec.~\ref{sec:supp_exp}), 
pseudo-code of the overall pipeline (Sec.~\ref{sec:supp_algo}), and implementation details (Sec.~\ref{sec:supp_imple}), 
% and discussion (Sec.~\ref{sec:supp_dis}), 
which are not presented in the main paper due to space limitations.

\begin{figure*}[t]
\centering
\resizebox{1\linewidth}{!}{%
\begin{tabular}{c}
\includegraphics[width=1\textwidth]{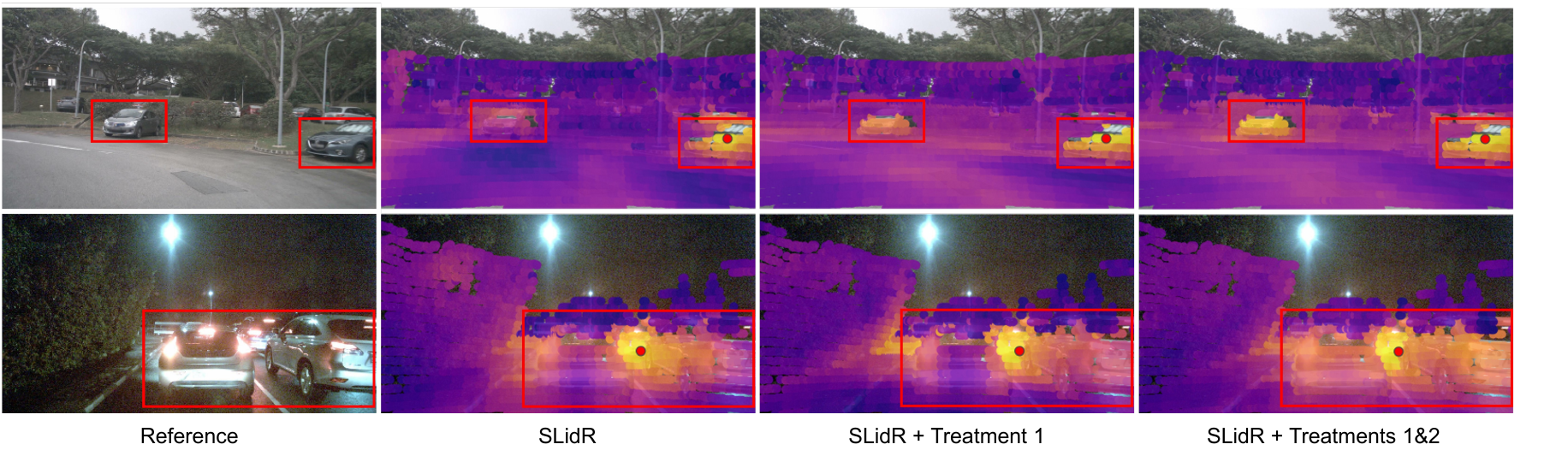}
\end{tabular}
}
\caption{
    \textbf{Qualitative results of cosine similarity. 
    } The cosine similarity between the query point (red dot) and the 3D point feature learned with SLidR, SLidR with treatment1, and SLidR with treatments 1 and 2. The projected points' colors go from violet to yellow for low and high similarity, respectively. We show these results in the validation set of nuScenes~\cite{caesar2020nuscenes}. This result shows that SLidR with treatments learns a more coherent 3D representation of the same objects.
    }
\label{figure:supp_fig1}
\end{figure*}

\section{Additional Qualitative Results}\label{sec:supp_qual}
\paragraph{Cosine Similarity} We present the 3D feature cosine similarity maps (See Fig.~\ref{figure:supp_fig1}).
After extracting 3D point-wise features through a pre-trained 3D model, cosine similarity is computed between the query point (red dot) feature and all the other point features.
Then, visualization is performed by projection to the corresponding image.
The projected points’ colors go from violet to yellow for low and high similarity, respectively.
The results show that a pre-trained model with treatments learns a more coherent 3D representation of the same objects.

\section{Additional Experiments}\label{sec:supp_exp}

\subsection{Keyframe Only}
As shown in Table 2 of the main paper, our method utilizes the unsynced inter-frame LiDAR point clouds from the nuScenes dataset~\cite{caesar2020nuscenes}. 
To ensure a fair comparison with previous image-to-LiDAR distillation methods that only use synced keyframe data, we report the results of our method using only the keyframe data. 
The Ours-Keyframe method matches keyframe images with keyframe LiDAR from different timestamps rather than matching inter-frame LiDAR with keyframe images to compose unsynced data (See~\Tref{table:keyframeonly}b).
Although relying solely on keyframes can increase the misalignment between points and pixels, our method still outperforms the previous methods (See~\Tref{table:keyframeonly}a).

\begin{table}[t]
    \caption{\textbf{3D semantic segmentation results on nuScenes and SemanticKITTI validation sets.} We compare our method using only the keyframe with the existing 3D representation learning methods using the nuScenes and SemanticKITTI datasets. Our method using only the keyframe surpasses the existing methods across all metrics.
    }
    \centering
    \begin{tabular}{c c c}
        \resizebox{0.5\linewidth}{!}{
            \begin{tabular}{m{3cm}ccc}
                \toprule
                \multirow{2}{*}{Method} & \multicolumn{2}{c}{nuScenes} & \multicolumn{1}{c}{SemanticKITTI} \\
                 \cmidrule(lr){2-3} \cmidrule(lr){4-4} 
                 & Lin. Prob. & 1$\%$ & 1$\%$ \\
                 \midrule
                 Random & 8.1 & 30.3 & 39.5 \\
                 PointContrast~\cite{xie2020pointcontrast} & 21.9 & 32.5~\textcolor{ballblue}{(+2.2)}  & 41.1~\textcolor{ballblue}{(+1.6)} \\ 
                 DepthContrast~\cite{zhang2021self} & 22.1 & 31.7~\textcolor{ballblue}{(+1.4)}  & 41.5~\textcolor{ballblue}{(+2.0)} \\ 
                 PPKT~\cite{liu2021learning} & 36.4 & 37.8~\textcolor{ballblue}{(+7.5)} & 43.9~\textcolor{ballblue}{(+4.4)} \\ 
                 SLidR~\cite{sautier2022image} & 38.8 & 38.2~\textcolor{ballblue}{(+7.9)} &  44.6~\textcolor{ballblue}{(+5.1)} \\ 
                 ST-SLidR~\cite{mahmoud2023self} & 40.4 & 40.7~\textcolor{ballblue}{(+10.4)} & 44.7~\textcolor{ballblue}{(+5.2)} \\ 
                 TriCC~\cite{pang2023unsupervised}  & 38.0 & 41.2~\textcolor{ballblue}{(+10.9)} & 45.9~\textcolor{ballblue}{(+6.4)} \\ 
                 \midrule
                
                 \textbf{Ours-Keyframe} & \textbf{46.3} & \textbf{41.6}~\textbf{\textcolor{ballblue}{(11.3)}} & \textbf{50.3}~\textbf{\textcolor{ballblue}{(10.8)}} \\ 
                
                \bottomrule
                \end{tabular}
        }
        & &
            \raisebox{-0.5\height}{\includegraphics[width=0.4\linewidth]{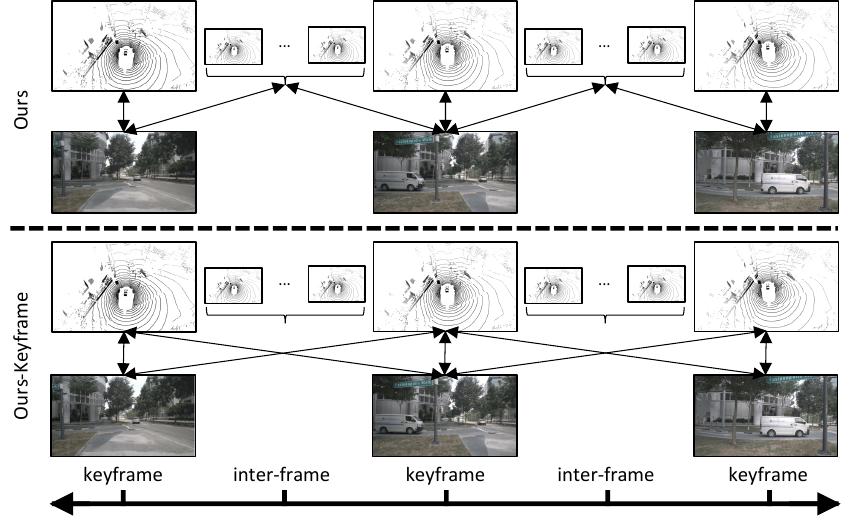}}
        \\
        \small{(a)}
        & &
        \small{(b)}
    \end{tabular}
    \label{table:keyframeonly}
\end{table}

\subsection{Various Voxel Sizes}
We report how changing the voxel size for each coordinate affects linear probing performance on the nuScenes dataset. We denote the voxel size for cylindrical coordinate as the values of $\delta\rho$ and $\delta z$.

In cylindrical coordinate, a voxel size of 10$cm$, and in Cartesian coordinate, a voxel size of 5$cm$, are found to be best. Regardless of the coordinate system, excessively reducing the voxel size to minimize quantization error can lead to a significant decrease in performance (See Table~\ref{table:coord_voxel}).
While over-reducing the voxel size decreases quantization error and increases the number of preserved raw points, it can result in sparse data, with most voxels being empty. 
This sparsity can pose challenges for 3D networks in effectively learning and recognizing patterns.

\begin{table}[t]
\caption{\textbf{Impact of coordinate and various voxel sizes.} We report that changing the voxel size for each coordinate affects linear probing performance on the nuScenes dataset, with optimal sizes being 10$cm$ in cylindrical coordinates and 5$cm$ in Cartesian coordinates. However, excessively reducing the voxel size results in data sparsity, which poses challenges for 3D network pattern recognition.
}
\centering
    \resizebox{0.5\linewidth}{!}{
    \begin{tabular}{|l|c|c|c|c|}
    \hline
    \diagbox{Coordinate}{Voxel Size ($cm$)} & 1 & 5 & 10 & 20 \\ \hline
    Cylindrical & 33.0 & 38.0 & 38.8 & 37.5 \\ \hline
    Cartesian & 31.3 & 41.2 & 40.8 & 40.8 \\ \hline
    \end{tabular}}
    \label{table:coord_voxel}
\end{table}

\subsection{The Number of Sampling Inter-frame LiDAR}
We report the performance variations based on the number of LiDAR samplings in inter-frame data when utilizing unsynced data. 
The difference in performance between sampling once and sampling twice is negligible (See~\Tref{table:syncedonly}).
This indicates that the utilization of unsynced data itself is crucial.

% \begin{table}[t]
% \caption{\textbf{Ablation study on the number of sampling inter-frame LiDAR.}
% Performance improvements are marginal whether sampling once or twice in linear probing tests. This indicates the importance of using unsynced data itself to achieve better results.}
% \centering
%     \resizebox{0.4\linewidth}{!}{
%     \begin{tabular}{cc}
%     \toprule
%     \multirow{2}{*}{\# inter-frame} & nuScenes \\ 
%     \cmidrule(lr){2-2}
%     & Lin Prob (100\%) \\
%     \midrule
%     Synced Only & 41.2 \\
%     1 & 45.2 \\
%     2 & 45.3 \\
%     \bottomrule
%     \end{tabular}}
%     \label{sample}
% \end{table}

\begin{table}[t]
    \caption{\textbf{Ablation study on the number of sampling inter-frame LiDAR.}
    Performance improvements are marginal whether sampling once or twice. This indicates the importance of using unsynced data itself to achieve better results.}
    \centering
    \begin{tabular}{c c c}
        \resizebox{0.5\linewidth}{!}{
            \begin{tabular}{cc}
                \toprule
                \multirow{2}{*}{\# inter-frame} & nuScenes \\ 
                \cmidrule(lr){2-2}
                & Lin. Prob. (100$\%$) \\
                \midrule
                Synced Only & 41.2 \\
                1 & 45.2 \\
                2 & 45.3 \\
                \bottomrule
            \end{tabular}
        }
        & &
            \raisebox{-0.5\height}{\includegraphics[width=0.4\linewidth]{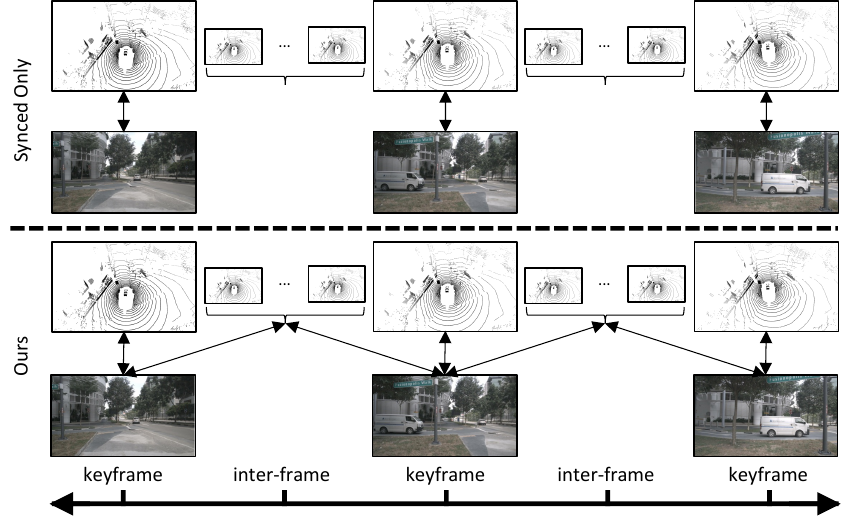}}
        \\
        \small{(a)}
        & &
        \small{(b)}
    \end{tabular}
    \label{table:syncedonly}
\end{table}

\subsection{Dynamic Point Cloud Accumulation}
We compare PPM's point cloud accumulation performance with that of existing methods, WsRSF~\cite{gojcic2021weakly} and PCAccumulation~\cite{huang2022dynamic}. PPM is based on an unsupervised method, WsRSF on a weakly supervised method, and PCAccumulation on a supervised method to point cloud accumulation.
PPM demonstrates good performance on static parts but has room for improvement in handling dynamic parts, which are a primary cause of misalignment on unsynced data (See~\Tref{sceneflow}).
As shown in Table 2 of the main paper, correcting misalignments with PPM can enhance the performance of image-to-LiDAR distillation. 
Following this trend, we expect that replacing PPM with a more effective point cloud accumulation method could further improve the performance of image-to-LiDAR distillation. 
However, LiDAR 3D scene flow needs to be trained whenever the data domain is changed; its model designs often cover a limited range and point cloud. Therefore, we propose the PPM, whose design motivation is its unsupervised manner and versatility with respect to LiDAR characteristics and environments.
\begin{table}[t]
    \caption{
    \textbf{Dynamic Point Cloud Accumulation results on nuScenes.}
Point cloud accumulation performances show that PPM, an unsupervised method, excels in static parts but has room for improvement with dynamic parts, in contrast to WsRSF, a weakly supervised method, and PCAccumulation, a supervised method.
    }
\centering
    \resizebox{1.0\linewidth}{!}{
    \begin{tabular}{m{2.5cm}cccccccccc}
    \toprule
    \multirow{2}{*}{Method}
    & \multirow{2}{*}{Strategy}
    & \multicolumn{4}{c}{Static part} & \multicolumn{5}{c}{Dynamic foreground} \\
    \cmidrule(lr){3-6}
    \cmidrule(lr){7-11}
    & & EPE avg.$\downarrow$ & AccS$\uparrow$ & AccR$\uparrow$ & ROutlier$\downarrow$ & EPE avg.$\downarrow$ & EPE med.$\downarrow$ & AccS$\uparrow$ & AccR$\uparrow$ & ROutliers$\downarrow$ \\
    \midrule
    N/A & - & 1.452 & 18.0 & 19.5 & 74.1 & 1.903 & 1.017 & 2.4 & 6.5 & 79.7 \\ 
    WsRSF~\cite{gojcic2021weakly} & Weakly & 0.195 & 57.4 & 82.6 & 4.8 & 0.539 & 0.204 & 17.9 & 37.4 & 32.0 \\ 
    PCAccumulation~\cite{huang2022dynamic} & Supervised & 0.111 & 65.4 & 88.6 & \textbf{1.1} & \textbf{0.301} & \textbf{0.146} & \textbf{26.6} & \textbf{53.4} & \textbf{12.1} \\ 
    PPM & Unsupervised & \textbf{0.102} & \textbf{83.0} & \textbf{89.2} & 5.0 & 0.992 & 0.409 & 13.3 & 26.3 & 49.5 \\
    \bottomrule
    \end{tabular}
                                  }\vspace{-3mm}
    \label{sceneflow}
\end{table}

\subsection{Additional Verification of Treatment1}
We verify the usefulness of treatment1 by adapting different voxel-based networks and conducting multiple runs.~\Tref{table:treatment1} shows that voxel-based networks with treatment1 consistently outperform those without it, even after multiple runs.
\begin{table}[!t] \caption{\textbf{Additional Verification of Treatment 1.} We report the performance of SLidR using different 3D voxel-based backbones, divided at the midline, with both Cylindrical and Cartesian coordinates. The results are presented for three different runs. Specifically, VoxelNet is pre-trained for 20 epochs. Consistent improvements are observed with the use of treatment1 across multiple runs and different backbones.}
    \vspace{-3mm}
    \centering
    \resizebox{\linewidth}{!}{
    \begin{tabular}{l@{\hskip 0.2in}c@{\hskip 0.2in}c@{\hskip 0.2in}c@{\hskip 0.2in}c@{\hskip 0.2in}c}
        \toprule
        \multirow{2}{*}{Method} & \multirow{2}{*}{Coordinate} & \multirow{2}{*}{3D backbone} &  & nuScenes (100$\%$) & \\
        \cmidrule(lr){4-6}
        & &  & Lin. Prob. (run 1) & Lin. Prob. (run 2)  & Lin. Prob. (run 3) \\ 
        \midrule SLidR & Cylindrical & MinkUNet & 38.8 & 38.4  & 38.9 \\
        SLidR & \textbf{Cartesian} & MinkUNet & \textbf{41.6} & \textbf{41.2} & \textbf{41.8} \\
        \midrule
        SLidR & Cylindrical & VoxelNet & 25.0 & 25.5 & 25.3 \\
        SLidR & \textbf{Cartesian} & VoxelNet & \textbf{27.3} & \textbf{26.8} & \textbf{26.3} \\
        \bottomrule
    \end{tabular}}
    \vspace{-3mm}
    \label{table:treatment1}
\end{table}

\subsection{Complexity and Extra Memories for Treatments}
Using PPM, the whole nuScenes dataset can be processed in 5 hours. As a data preprocessing method, PPM only needs to be performed once.~\Tref{table:resource} shows per GPU memory consumption, per epoch training time, and linear probing performance (LP) when applying Treatment 1\&2. While our treatments slightly increase memory consumption and training time, our performance improvement is notable compared to (E) SLidR with similar resource
\begin{table}[!t]
    \caption{\textbf{Resources required for Treatment 1\&2.} We report the per GPU memory consumption, per epoch training time, and linear probing performance (LP) when applying Treatment 1\&2. While our treatments slightly increase memory consumption and training time, performance improvement is notable compared to SLidR, which has similar resources.}
    \vspace{-3mm}
    \centering
    \resizebox{\linewidth}{!}{
    \begin{tabular}{c@{\hskip 0.2in}l@{\hskip 0.2in}c@{\hskip 0.2in}c@{\hskip 0.2in}c@{\hskip 0.1in}c@{\hskip 0.1in}c@{\hskip 0.1in}}
        \toprule
         & Method & Epoch & Batch size & Memory [MB] & Time [hour] & LP \\
        \midrule
        (A) & SLidR (Cylindrical) & 50 & 16 & 10.4 & 0.5 & 38.8 \\
        \midrule
        (B) & + Treatment 1 (Cartesian) & 50 & 16 & 12.8 & 0.7 & 41.2 \\
        (C) & + Treatment 2 (PPM) & 50 & 32 & 18.4 & 0.9 & 41.2 \\
        \midrule
        (D) & + Treatment 1\&2 (ours) & 50 & 32 &21.6 & 1.2 & 45.2 \\
        (E) & + Treatment 1\&2 (ours) & 20 & 16 &13.0 & 0.5 & 44.7 \\
        \bottomrule
    \end{tabular}}
    \vspace{-3mm}
    \label{table:resource}
\end{table}

\subsection{Different 2D Backbone for Distillation}
To verify that our treatments are consistently applicable to different pre-trained 2D backbones, we replace the pre-trained 2D backbone with a ViT-S/8 model trained with DINO and report the performance.~\Tref{table:2d_backbone} shows consistent performance improvements, demonstrating that our treatments are effective across various 2D backbones.

\begin{table}[!t]
    \caption{\textbf{Linear Probing results of different 2D backbones.} We report the performance of various methods using different pre-trained 2D backbones, specifically comparing MoCov2 and DINOv1. The methods are pre-trained on the nuScenes dataset. The results demonstrate that our treatments show the highest improvement in performance across both backbones, indicating consistent applicability of our treatments to various 2D backbones.}
    \vspace{-3mm}
    \centering
    \resizebox{0.5\linewidth}{!}{
    \begin{tabular}{lccc}
        \toprule
         \multirow{2}{*}{Method} & Pretrain & \multicolumn{2}{c}{2D backbone} \\
         \cmidrule(lr){3-4}
         & dataset & MoCov2 & DINOv1 \\
         \midrule
         % PPKT & MoCov2 & 36.4 \\
         % SLidR & MoCov2 & 38.8 \\
         % Ours & MoCov2 & \textbf{45.2} \\
         % \midrule
         PPKT & Nuscenes & 36.4 & 38.6 \\
         SLidR & Nuscenes & 38.8 & 39.3 \\
         Ours & Nuscenes & \textbf{45.2} & \textbf{47.3} \\
        \bottomrule
    \end{tabular}}
    \vspace{-3mm}
    \label{table:2d_backbone}
\end{table}

\subsection{Results of Different Frame Gaps}
We experiment to see how performance changes according to different frame gaps from keyframe data. We evaluate our treatments using a ViT-S/8 2D backbone trained with DINO, measuring LP performance with frame gaps up to 40.
~\Tref{table:frame_gap} shows that performance increases as the frame gap increases up to 10 frames and then decreases as the frame gap continues to increase. The performance increase is likely due to data diversity, while the decrease is likely due to the expected increase in PPM errors.

\begin{table}[!t] \caption{\textbf{Linear Probing results of different frame gaps.} We report the linear probing (LP) performance for different frame gaps, ranging from 1 to 40. The results indicate that performance increases as the frame gap increases up to 10 frames and then slightly decreases as the frame gap continues to increase.}
    \vspace{-3mm}
    \centering
    \resizebox{\linewidth}{!}{
    \begin{tabular}{lcccccc}
        \toprule
        \multirow{2}{*}{Method} & \multicolumn{6}{c}{Frame Gap} \\
        \cmidrule(lr){2-7}
        & \hspace{5mm}1\hspace{5mm} & \hspace{5mm}5\hspace{5mm} & \hspace{5mm}10\hspace{5mm} & \hspace{5mm}20\hspace{5mm} & \hspace{5mm}30\hspace{5mm} & \hspace{5mm}40\hspace{5mm} \\  
        \midrule Ours & 45.6 & 47.3 & 47.3 & 47.1  & 47.2 & 46.5 \\
        \bottomrule
    \end{tabular}}
    \vspace{-3mm}
    \label{table:frame_gap}
\end{table}

\section{Algorithm}\label{sec:supp_algo}
In this section, we provide a pseudo-code of our overall pipeline in Algo.\ref{algo:pseudo_code}. 
The \textit{sampling} function is designed to alleviate data redundancy by more frequently sampling data from inter-frames that are farthest from the keyframe.
Transformation Z is initialized to an identity matrix for all points, and the computed transformation is assigned only to points classified as moving.
Within the \textit{transformation} function, x\_p\_s
is transformed to global coordinates, transformed by Z, and restored to LiDAR sensor coordinates.
Within the \textit{pixelPointMatching} function, x\_p\_t and x\_p\_s\_trans are transformed from LiDAR sensor coordinates to camera sensor coordinates and projected to the 2D coordinate of x\_i\_t through the camera intrinsic matrix.
The \textit{pixelPointMatching} function creating a pixel-point matching index corresponds to function $T$, first mentioned in Sec. 3.2 of the main paper.
For brevity, the algorithm does not include the matching of SLIC~\cite{achanta2012slic} based superpixels and corresponding point clouds matching.

\begin{figure*}[t]
\centering
\resizebox{0.8\linewidth}{!}{
\begin{algorithm}[H]
    \SetAlgoLined
        \PyComment{f\_p, f\_i: 3d and 2d network} \\
        \PyComment{x\_p\_t, x\_p\_s: 3d point cloud at keyframe t and inter-frame s} \\
        \PyComment{x\_p\_s\_list: list of 3d point cloud at inter-frames} \\
        \PyComment{x\_i\_t: 2d image at keyframe t} \\
        % \PyComment{$\tau$: \text{temperature}} \\
        \PyComment{aug\_p, aug\_i: Augmentations for point clouds and images} \\
        \PyComment{quant: Quantization for point clouds} \\
        \PyCode{} \\
        \PyCode{for x\_p\_t, x\_p\_s\_list, x\_i\_t in loader:} \\
        \Indp
        \PyCode{} \\
        \PyCode{Z = ppm(x\_p\_t, x\_p\_s\_list, x\_i\_t)}\PyComment{get transformation Z} \\
        \PyCode{} \\
        \PyCode{s\_i = sampling(x\_p\_s\_list)}\PyComment{get index at inter-frame s} \\
        \PyCode{x\_p\_s, Z = x\_p\_s\_list[s\_i]}, Z[s\_i] \\
        \PyCode{x\_p\_s\_trans = transformation(x\_p\_s, Z)}\PyComment{get transformed points} \\
        \PyCode{} \\
        \PyComment{get positive pixel-point matching index} \\
        \PyCode{i\_t\_i, p\_t\_i = pixelPointMatching(x\_i\_t, x\_p\_t)} \\
        \PyCode{i\_s\_i, p\_s\_i = pixelPointMatching(x\_i\_t, x\_p\_s\_trans)} \\
        \PyCode{} \\
        \PyComment{augment, quantize, and feed-forward} \\
        \PyCode{F\_p\_t, F\_p\_s, F\_i\_t = f\_p(quant(aug\_p(x\_p\_t))), f\_p(quant(aug\_p(x\_p\_s))), f\_i(aug\_i(x\_i\_t))} \\
        \PyCode{} \\
        \PyComment{pair point and pixel-wise feature} \\
        \PyCode{F\_p\_t, F\_p\_s, F\_i\_t, F\_i\_s = F\_p\_t[p\_t\_i], F\_p\_s[p\_s\_i], F\_i\_t[i\_t\_i], F\_i\_t[i\_s\_i]} \\
        \PyCode{} \\
        \PyCode{loss = contrastDistill(cat(F\_p\_t, F\_p\_s), cat(F\_i\_t, F\_i\_s))} \\
        \PyCode{loss.backward()} \\
        \PyCode{update(f\_p, f\_i)} \\
        \PyCode{} \\
        \Indm
        \PyCode{def ppm(x\_p\_t, x\_p\_s\_list, x\_i\_t):} \\
        \Indp
        \PyCode{} \\
        \PyCode{Z = eye(4).reshape((1, 4, 4)).repeat(len(cat(x\_p\_t, x\_p\_s\_list)), 1, 1)} \\
        \PyCode{} \\
        \PyComment{aggregation in the global coordinate} \\
        \PyCode{x\_p\_t, x\_p\_s\_list = sensor2global(x\_p\_t), sensor2global(x\_p\_s\_list)} \\
        \PyCode{x\_p = cat(x\_p\_t, x\_p\_s\_list)} \\
        \PyCode{} \\
        \PyCode{g\_i = groundRemoval(x\_p)}\PyComment{get ground point index} \\
        \PyCode{x\_p\_ng = x\_p[$\sim$ \text{g\_i}]}\text{\PyComment{get non-ground point}} \\
        \PyCode{} \\
        \PyCode{c\_i = clustering(x\_p\_ng)}\PyComment{get cluster index} \\
        \PyCode{} \\
        \PyCode{m\_i = movingClusterTracking(x\_p\_ng, c\_i)}\PyComment{get moving point index} \\
        \PyCode{} \\
        \PyCode{x\_p\_m, c\_i = x\_p\_ng[m\_i], c\_i[m\_i]}\PyComment{get moving point and moving cluster index} \\
        \PyCode{Z[$\sim$\text{g\_i][m\_i] = clusterWiseICP(x\_p\_m, c\_i)}}\text{\PyComment{assign computed transformation}} \\
        \PyCode{} \\
        \PyCode{return Z[len(x\_p\_t):]}\PyComment{return transformation Z at inter-frames} \\
        \Indm
        \PyCode{} \\
        \PyCode{def contrastDistill(F\_p, F\_i):} \\
        \Indp
        \PyCode{} \\
        \PyCode{logits = mm(norm(F\_p), norm(F\_i).T)} \\
        \PyCode{loss = crossEntropyLoss(logits/$\tau$ \text{, range(len(F\_p)))}} \\
        \PyCode{} \\
        \PyCode{return loss} \\
        \Indm
    \caption{PyTorch-style pseudo-code of ours overall pipeline.}
    \label{algo:pseudo_code}
\end{algorithm}
 }
\end{figure*}

\section{Implementation Details}\label{sec:supp_imple}
\begin{figure*}[t]
\centering
\resizebox{1\linewidth}{!}{%
\begin{tabular}{c}
\includegraphics[width=1\textwidth]{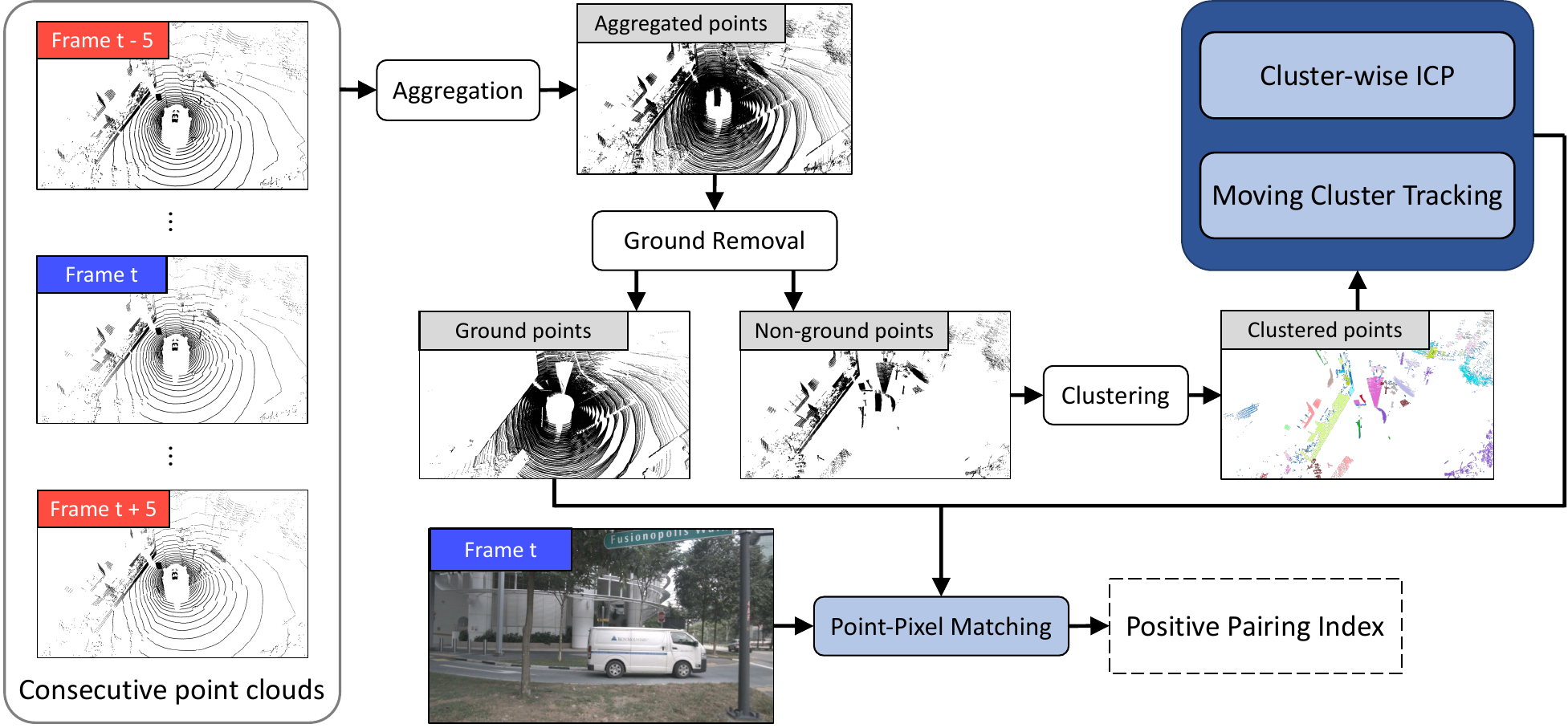}
\end{tabular}
}
\caption{
    \textbf{The overview of the Positive Pair Mining (PPM) module.
    } The Positive Pair Mining consists of four components, \ie,
    aggregation, ground removal, clustering, moving cluster tracking, cluster-wise ICP, and point-pixel matching steps. The aggregation step aggregates consecutive point clouds in the global coordinate. The ground removal step separates aggregated points into ground and non-ground points. The moving cluster tracking and cluster-wise ICP transform all the moving points from the inter-frames into the nearest keyframe $t$. The point-pixel matching step constructs positive pairs of 3D-2D by projecting transformed 3D points to the 2D image at keyframe $t$.
}
\label{figure:3}
\end{figure*}

\begin{figure}[t]
\centering
\resizebox{1\linewidth}{!}{%
\begin{tabular}{c}
\includegraphics[width=1\textwidth]{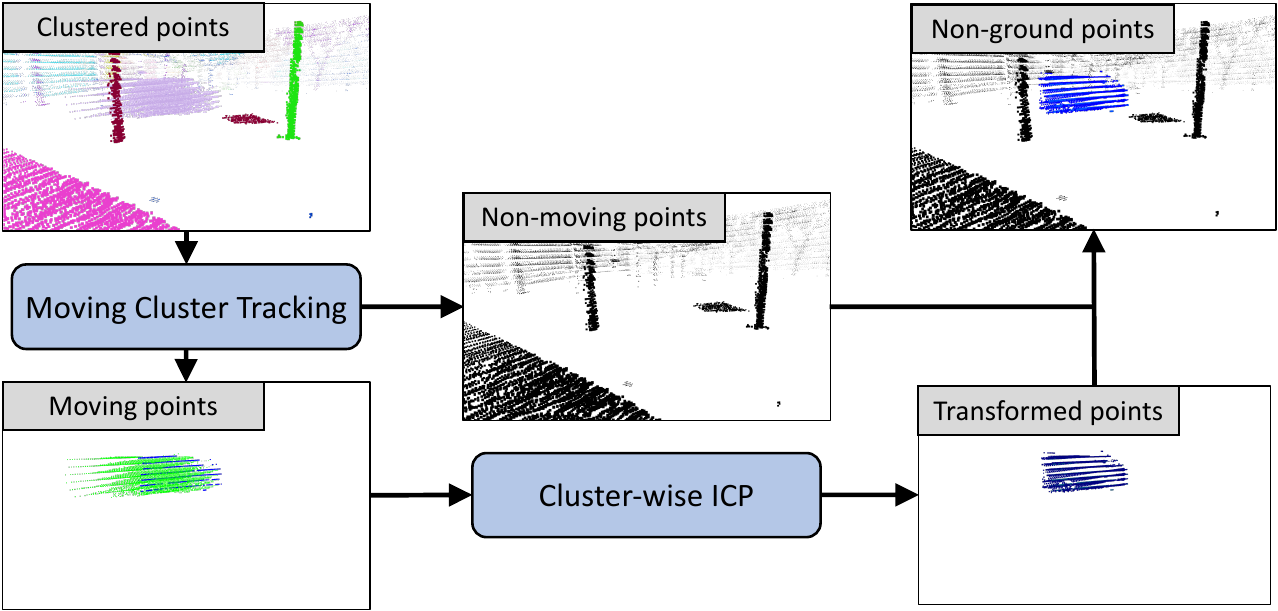}
\end{tabular}
}
\caption{
    \textbf{The pipeline of moving cluster tracking and cluster-wise ICP}. (moving cluster tracking) To distinguish the non-moving and moving points from the clustered points, we form each cluster’s points in consecutive times and then measure their center coordinates.
    If any $L1$-distance is larger than the threshold $c$, we categorize the points in the cluster into moving points.
(cluster-wise ICP) Using an unsupervised point cloud matching method, we obtain the 3D transformation $Z$ to the keyframe for generating the positive pairing index.
}
\label{figure:4}
\end{figure}

\paragraph{Positive Pair Mining (PPM)}\label{sec:supp_details}
Figure~\ref{figure:3} describes the overall scheme of PPM. The $11$ number of consecutive point clouds $\{P^{t - 5}, ...,  P^{t}, ..., P^{t + 5}\}$ are aggregated in the global coordinate through the relative poses readily obtained from GPS and IMU~\cite{geiger2012we}. We first split the aggregated points into ground and non-ground points in sequence using an unsupervised ground removal method~\cite{lee2022patchwork++}. The non-ground points are converted to clustered points using HDBSCAN~\cite{campello2013density}, and they pass through two consecutive steps: \textit{Moving cluster tracking} and \textit{Cluster-wise ICP}, as shown in~\Fref{figure:4}.
Moving cluster tracking identifies clusters that are in motion.
We form each cluster's points in consecutive times and then calculate their center coordinates. 
If any $l_{1}$ distance between the center coordinates of consecutive times exceeds the threshold $c$, we categorize the points in the cluster as moving points and non-moving ones otherwise. We set $c$ to $0.5$ meter. 
The clusters of moving points are fed to the Cluster-wise ICP. In the Cluster-wise ICP, we apply an unsupervised point cloud matching~\cite{rusinkiewicz2001efficient} to each moving cluster by exploiting the keyframe as the reference frame. This mining process outputs the 3D transformation $Z$ for each cluster that is combined with $T$ to obtain the positive pixel-point matching index.

For ground removal, we utilize the patchwork++\footnote{\postechred{https://github.com/url-kaist/patchwork-plusplus}}~\cite{lee2022patchwork++}.
Because the official implementation is designed for the SemanticKITTI dataset~\cite{behley2019semantickitti}, to apply it to the nuScenes dataset~\cite{caesar2020nuscenes}, 
we modified to set the mean coordinates of the aggregated point clouds to zero by subtracting the mean coordinates.
For the HDBSCAN clustering~\cite{campello2013density}, we set a minimum number of clusters to $50$, the number of clusters to $300$, $\alpha$ to 1, distance metric to Euclidean, and the leaf size to 100.
For the mean tracking in cluster tracking.

\end{document}